\documentclass[final,3p,times]{elsarticle}

\usepackage{amssymb}
\usepackage{multirow}
\usepackage{url}
\usepackage{hyperref}
\usepackage{floatrow}
\usepackage{subcaption}
\journal{Expert Systems with Applications}

\begin{document}

\begin{frontmatter}

%% Title, authors and addresses

%% use the tnoteref command within \title for footnotes;
%% use the tnotetext command for theassociated footnote;
%% use the fnref command within \author or \address for footnotes;
%% use the fntext command for theassociated footnote;
%% use the corref command within \author for corresponding author footnotes;
%% use the cortext command for theassociated footnote;
%% use the ead command for the email address,
%% and the form \ead[url] for the home page:
%% \title{Title\tnoteref{label1}}
%% \tnotetext[label1]{}
%% \author{Name\corref{cor1}\fnref{label2}}
%% \ead{email address}
%% \ead[url]{home page}
%% \fntext[label2]{}
%% \cortext[cor1]{}
%% \affiliation{organization={},
%%             addressline={},
%%             city={},
%%             postcode={},
%%             state={},
%%             country={}}
%% \fntext[label3]{}

%\title{}

%% use optional labels to link authors explicitly to addresses:
%% \author[label1,label2]{}
%% \affiliation[label1]{organization={},
%%             addressline={},
%%             city={},
%%             postcode={},
%%             state={},
%%             country={}}
%%
%% \affiliation[label2]{organization={},
%%             addressline={},
%%             city={},
%%             postcode={},
%%             state={},
%%             country={}}

\title{PURSUhInT: In Search of Informative Hint Points Based on Layer Clustering for Knowledge Distillation}
 \author{Reyhan Kevser Keser\corref{cor1}\fnref{label_be}}
 \ead{keserr@itu.edu.tr}
% \ead[url]{home page}
  \author{Aydin Ayanzadeh\fnref{label_umbc}}
  \ead{aydina1@umbc.edu}
  \author{Omid Abdollahi Aghdam\fnref{label_arc}}
  \ead{abdollahi15@itu.edu.tr}
  \author{Caglar Kilcioglu\fnref{label_arc}}
  \ead{caglar.kilcioglu@arcelik.com}
  \author{Behcet Ugur Toreyin\fnref{label_be}\corref{eq_sen_}}
  \ead{toreyin@itu.edu.tr}
  \author{Nazim Kemal Ure\fnref{label_ai}\corref{eq_sen_}}
  \ead{ure@itu.edu.tr}

 \cortext[cor1]{Corresponding author}
 \fntext[label_be]{Signal Processing for Computational Intelligence (SP4CING) Research Group, Informatics Institute, Istanbul Technical University, Turkey}
 \fntext[label_umbc]{Department of Computer Science and Electrical Engineering, University of Maryland, Baltimore County, United States%, Baltimore, MD 21250, United States
 }
 \fntext[label_arc]{Arcelik Research and Development, Turkey}
\fntext[label_ai]{Artificial Intelligence and Data Science Application and Research Center, Istanbul Technical University, Turkey}
\cortext[eq_sen_]{Equal Senior Contribution}

\tnotetext[]{Full-text published in Expert Systems with Applications is available at \url{https://doi.org/10.1016/j.eswa.2022.119040} .}
\tnotetext[]{This work is licensed under a \href{http://creativecommons.org/licenses/by-nc-nd/4.0/}{Creative Commons Attribution-NonCommercial-NoDerivatives 4.0 International License}.}

\newpageafter{author}
\begin{abstract}
%% Text of abstract
One of the most efficient methods for model compression is hint distillation, where the student model is injected with information (hints) from several different layers of the teacher model. Although the selection of hint points can drastically alter the compression performance, conventional distillation approaches overlook this fact and use the same hint points as in the early studies. Therefore, we propose a clustering based hint selection methodology, where the layers of teacher model are clustered with respect to several metrics and the cluster centers are used as the hint points. Our method is applicable for any student network, once it is applied on a chosen teacher network. The proposed approach is validated in CIFAR-100 and ImageNet datasets, using various teacher-student pairs and numerous hint distillation methods. Our results show that hint points selected by our algorithm results in superior compression performance compared to state-of-the-art knowledge distillation algorithms on the same student models and datasets.

\end{abstract}

%%%%%%%%%%%%%%%%%%%%%%%%%%%

%%Graphical abstract

%\begin{graphicalabstract}
%		\centering
%		\includegraphics[width = 1.0\linewidth]{all_steps2.pdf}

%\end{graphicalabstract}
%%%%%%%%%%%%%%%%%%%%%%%%%%%

%%Research highlights-değişti bunlar:
%\begin{highlights}
%\item   We demonstrate that selection of hint points impacts the performance of knowledge distillation.
%\item	We propose a clustering based hint selection method.
%\item	Our method can be combined with any type of hint distillation that uses at least two hints.
%\item Our method is applicable for any student, once it is applied on a chosen teacher.
%%%Proposed method is applicable for any student model and valid for the whole training process, once it is applied on a chosen teacher model. 

%\end{highlights}
%%%%%%%%%%%%%%%%%%%%%%%%%%%%%%%%%%%

\begin{keyword}
%% keywords here, in the form: keyword \sep keyword
Model compression \sep Knowledge distillation \sep Hint selection \sep Feature matching

%% PACS codes here, in the form: \PACS code \sep code

%% MSC codes here, in the form: \MSC code \sep code
%% or \MSC[2008] code \sep code (2000 is the default)

\end{keyword}

\end{frontmatter}

%% \linenumbers

%% main text
\section{Introduction}
\label{intro}
Recent years have witnessed a tremendous increase in utilizing Deep Neural Networks (DNNs) for many computer vision tasks, including face recognition~\citep{facerecognition}, object detection \citep{yolov3}, classification \citep{alexnet,resnet}, segmentation \citep{seg2012}, tracking \citep{tracking2020} and self-driving cars \citep{self_driving2018}. That being said, many applications require these models to be implemented on edge devices~\citep{sze2017efficient}, and the sheer size and computation requirements of large-scale deep learning models yield significant performance drops when they are implemented on an edge device. The main objective of this paper is to push the state-of-the-art in compressing large scale neural networks to increase their inference speed without sacrificing too much accuracy. We present a novel knowledge distillation approach that extends ideas from unsupervised clustering and hint distillation that obtains state-of-the-art performance in compression of various teacher models.

The development of deep learning based computer vision models were fueled with the advent of the ILSVRC challenge~\citep{ILSVRC15}. Starting with AlexNet architecture~\citep{alexnet}, model performance is improved at the expense of significant increase in model size and depth with the introduction of VGGNet~\citep{VGG} and ResNet~\citep{resnet} over the years. Hence, it is widely known that there is a trade-off between model size and generalization performance in these models. Parallel to these developments, there has also been considerable research in model compression~\citep{choudhary2020comprehensive}, 
where several different methodologies were offered for converting large models into smaller ones while conserving the performance as much as possible. Among these approaches, knowledge distillation \citep{hinton2015distilling} is a popular choice, since the compressed network architectures (referred to as the student network) can be set according to performance limitations of the hardware it is going to be implemented on, which is a critical requirement for deploying machine learning models on edge devices. In particular, hint distillation methods \citep{romero2015fitnets, Zagoruyko2017AT, huang2022feature, li2022distilling} have been popular in compressing large scale teacher networks to smaller scale student networks. The success of these methods stem from injecting hint points into the compression process, so that student model is forced to match, not only the input-output response, but also the outputs of intermediate layers of the teacher network. That being said, optimizing the model compression performance in hint based distillation is still an open research area. 

The main contribution of this paper is a novel hint based knowledge distillation methodology, that attempts to optimize the location of hint points based on the information provided by clustering of teacher network layers according to different metrics. Our contributions can be summarized as follows:
\begin{itemize}
  \item We demonstrate that selection of hint points impacts the performance of the student network and using heuristics or rule-of-thumbs for hint locations may hurt the generalization accuracy.
 
  \item We propose a novel method to search for informative hint positions, where the main idea is to cluster layers of the teacher network and use cluster centers as hint positions. This is the first method that proposes a hint position selection technique based on layer clustering for hint based knowledge distillation purposes.

  \item Our method can be combined with any type of hint distillation that uses at least two hints. We demonstrate this property by conducting experiments for various hint distillation methods and numerous architectures. 

  \item Proposed method is applicable for any student model and valid for the whole training process, once it is applied on a chosen teacher model. 

  \item We compare our approach with the current state-of-the-art algorithms on CIFAR-100 and ImageNet datasets. Results show that our method yields superior results for compression of various teacher models regardless of having the same or different architectures with student models, on both of these datasets.

\end{itemize}

The rest of the paper is organized as follows. In Section~\ref{sec:related}, we present a comprehensive survey of the literature related to the scope of our study. In Section~\ref{sec:method}, the proposed method is explained in detail and the elaborated solution is introduced to tackle the problem. Experiments and results are reported in Section~\ref{sec:experiments} to assess the performance of the proposed approach. We conclude by presenting a discussion and suggesting future research directions. 

\section{Related work}
\label{sec:related}

Literature on model compression, knowledge and hint distillation are reviewed in this section.  

\subsection{Model Compression}

Recently, there has been a significant progress in the field of computer vision by leveraging DNNs. However, many  state-of-the-art models have failed to meet the requirements for real-world applications due to model's computational complexity and memory demand. Various model compression approaches, targeted at compressing DNNs, have been proposed to mitigate this issue. In general, model compression studies can be categorized as follows: the parameter pruning and quantization, low-rank factorization, compact convolutional filters and knowledge distillation~\citep{cheng2017survey}. 

In the parameter pruning techniques~\citep{deepcompression2016iclr}, removing redundant weights based on thresholding is explored. On the other hand, quantization based techniques represent parameters with less number of bits. Although, network pruning reduces model size, it does not usually decrease the inference time, which is critical for real-world applications on the edge devices. As for low-rank factorization, it is not easy to implement and requires expensive computation because of the decomposition operation. 

Compact convolutional filters are proposed in~\citep{szegedy2017inception} and adapted by SqueezeNet ~\citep{iandola2016squeezenet} and MobileNet~\citep{sandler2018mobilenetv2}. Methods using compact convolutional filters are generally computationally efficient, however, significantly decreased number of parameters in them might degrade the performance of the models in relatively complex tasks~\citep{howard2019searching}. Thus, we mainly focus on knowledge distillation, in this work.

\subsection{Knowledge Distillation}

Knowledge distillation (KD) is the process of transferring knowledge between networks, where one usually aims to transfer the knowledge of a big network (teacher) to a smaller/more compact network (student). KD is mostly known due to Hinton's work~\citep{hinton2015distilling}, while it was first proposed by~\citep{bucilua2006model}. The most well known form of KD uses the combination of soft targets produced by the teacher model and labels as the target in its objective function. The methods that uses soft targets without hints, are known as output (logit) distillation~\citep{heo2019comprehensive}. 

Soft targets are calculated as in~(\ref{eq:kd}), by passing training batches through teacher model and using softmax output layer with a higher temperature value $(T>1)$:
\begin{equation}
\label{eq:kd}
    p_{i} = \frac{exp(z_i/T)}{\sum_{j}{exp(z_j/T)}}
\end{equation}
where $p$, $z$ and $T$ show the softened class probability, logit and temperature values, respectively. For instance,  $i,j \in \{0, ..., 99\}$ for a classification task with $100$ classes, such as CIFAR-100. The same temperature value is used for generating both softened logits in teacher and student models, and higher $T$ results in a softer class probability distribution. The loss function to train the student network is evaluated in~(\ref{eq:L_Student}) as follows:
\begin{equation}
\label{eq:L_Student}
    L = \lambda {L_{logit}}(p^S, p^T) + (1-\lambda)L_{cls}(p^S, y)
\end{equation}
where $\lambda$ is the trade-off between {logit} distillation loss {$(L_{logit})$} and classification loss $(L_{cls})$, $y$ represents the label, $p^S$ and $p^T$ are the softened logits of student and teacher networks, respectively. Temperature is set to 1 in $L_{cls}$. Cross-entropy and Kullback-Leibler losses are generaly used for $L_{cls}$ and {$L_{logit}$}, respectively.

A recent study \citep{guo2020spherical} improves KD, where it introduces Spherical Knowledge Distillation (SKD) that projects all logits of the teacher and the student on a sphere by normalization. In SKD, logits are scaled into a unit vector and multiplied with the average teacher norm in order to recover its norm to the original level. \citep{zhao2022decoupled} introduces a novel distillation approach by reformulating the classical logit distillation. Furthermore, \citep{ren2022co} proposes a logit distillation approach in order to train a transformer student using lightweight teacher models with different architectures.

Logit distillation is an open research area, where studies are focused on different aspects such as data-free distillation~\citep{kang2021data}, online distillation \citep{lan2018knowledge},
ensemble distillation \citep{nam2022improving}
or federated learning \citep{ni2022federated}. On the other hand, hint distillation aims to provide more information to student by utilizing knowledge of intermediate layers of the teacher model~\citep{romero2015fitnets,Zagoruyko2017AT,heo2019comprehensive, huang2022feature}, which is also called as feature matching and constitutes the foundation of our proposed method. It is more advantageous compared to the logit distillation, due to allowing elimination of capacity gap between teacher and student models and applicability on wide range of problems such as regression and low-level vision problems \citep{wang2021knowledge}.

\subsection{Hint Distillation}

In KD~\citep{hinton2015distilling} and SKD~\citep{guo2020spherical}, only logits of teacher network are used as the soft target to train student network. Recently, methods that leverage intermediate feature maps are proposed to improve logit (output) distillation's performance~{\citep{romero2015fitnets, ahn2019variational, ji2021show, li2022distilling, zhang2021student, wu2022single}}.

In~\citep{romero2015fitnets} the output from middle layer of teacher network is chosen as a hint to guide the middle layer of the student network. This approach is called FitNet and involves a two-stage training strategy. 
%First, a pre-trained teacher network is used to train student network up to the guided layer, secondly, and then the classical KD approach is used for training the whole student network.
{In the first stage, the student network is trained up to the guided layer by the pre-trained teacher network. Then, logit distillation is used for the training of the whole student model, in the second stage.}
In~\citep{Zagoruyko2017AT} activation-based and gradient-based Attention Transfer (AT) is proposed to transfer knowledge from cumbersome teacher network to a smaller student network. In knowledge distillation using AT, the distance between $l_2$-normalized attention maps of teacher-student layer pairs are added to the standard cross entropy loss. In Variational Information Distillation (VID)~\citep{ahn2019variational}, the mutual information of the teacher-student layer pairs are maximized to transfer knowledge from teacher to student. In the classical KD~\citep{hinton2015distilling} softened class probabilities of teacher network are used in combination of cross entropy loss to train student network, and in FitNet~\citep{romero2015fitnets}, the middle layer of the teacher is also used to guide the middle layer of student during training. Furthermore, \citep{li2022distilling} applies hint distillation by combining self distillation and online distillation schemes. \citep{huang2022feature} proposes a hint distillation approach by addressing the issue of feature maps with different sizes that obtained from teacher and student models for the task of low resolution object recognition.
\citep{zhang2021self} introduces a self distillation scheme where the deepest layer transfers its knowledge to shallower layers in a model which is a multi-exit neural network.
In~\citep{heo2019comprehensive}, authors propose to transfer knowledge from before the activation function as in~\citep{heo2019knowledge}. Moreover, they propose employing a loss function and teacher transform, which are compatible with the feature position. Their study modifies the feature position, but keeps the conventional grouping approach. 

In IAKD \citep{fu2021interactive} which is a study that refers to the hint position, networks are divided into smaller parts compared to conventional approach. In that study, it is reported that using all break points for hint transfer results in worse performance than conventional grouping. In~\citep{ruffy2019state}, results pertained to experiments with various hint positions are mentioned to fail performing well. 

Recently, some studies focus on hint position problem, which is named as feature linking in some studies \citep{ji2021show}. For example, \citep{jang2019learning} propose a meta-learning scheme for transfer learning to determine hint positions as well as the weights for hints. However, this approach is costly and considers only last layers of blocks of a network as potential hint positions. \citep{ji2021show} presents an attention based approach which is called as AFD, for the same purpose. Nevertheless, the approach presented in \citep{ji2021show} repeats the search process for every iteration and is hard to implement for every potential hint positions for large networks such as resnet110. On the other hand, our approach is applicable for any student network and valid for the whole training process, once it is applied on a chosen teacher network. 

Furthermore, \citep{chen2021distilling} firstly proposed to transfer knowledge between different stages of teacher and student models. However, their approach relies on the conventional grouping of network layers. \citep{sun2019patient} suggests two schemes for hint positions on several problems of natural language processing, which are choosing every $k^{th}$ layers and choosing last $k$ layers as the hint positions. It should be noted that their approach employs a fixed rule for hint positions without considering the information that layers hold. \citep{deng2022distpro} employs a meta-learning scheme to choose hint positions and hint transformations to align them between models. Nevertheless, their search space is limited by the last layers of conventional groups for hint positions. On the other hand, our approach re-configures the grouping of layers considering the knowledge that layers contain, which allows all layers to be chosen as hint positions.
%in order to avoid transferring redundant information during distillation.
On a similar account,~\citep{haidar2021rail}  suggests utilizing a non-conventional, random-selection-based hint position strategy for distillation.~\citep{li2022knowledge} focuses on eliminating the redundant knowledge in the distillation process, akin to our approach. Their method makes a selection among the samples whose knowledge is being transferred. However, our approach operates on the hint positions at the teacher network. 
\citep{passban2021alp} utilizes attention based approach, where they focus on natural language models. Their method mainly fuses knowledge from different layers of the teacher for transfer.

To sum up, conventional hint distillation studies in the literature, obtain intermediate feature maps to transfer teacher's knowledge by grouping layers of teacher network. This is accomplished by splitting layers according to the spatial sizes of the feature maps. Contrary to the strategy of following a pre-determined approach for hint position selection adopted by the state-of-the-art knowledge distillation methods, in this work, we propose a layer clustering method as a systematic alternative to choose more informative hint points.

\section{Method}
\label{sec:method}

\begin{figure}[H]
		\centering
		\includegraphics[width = 1.0\linewidth]{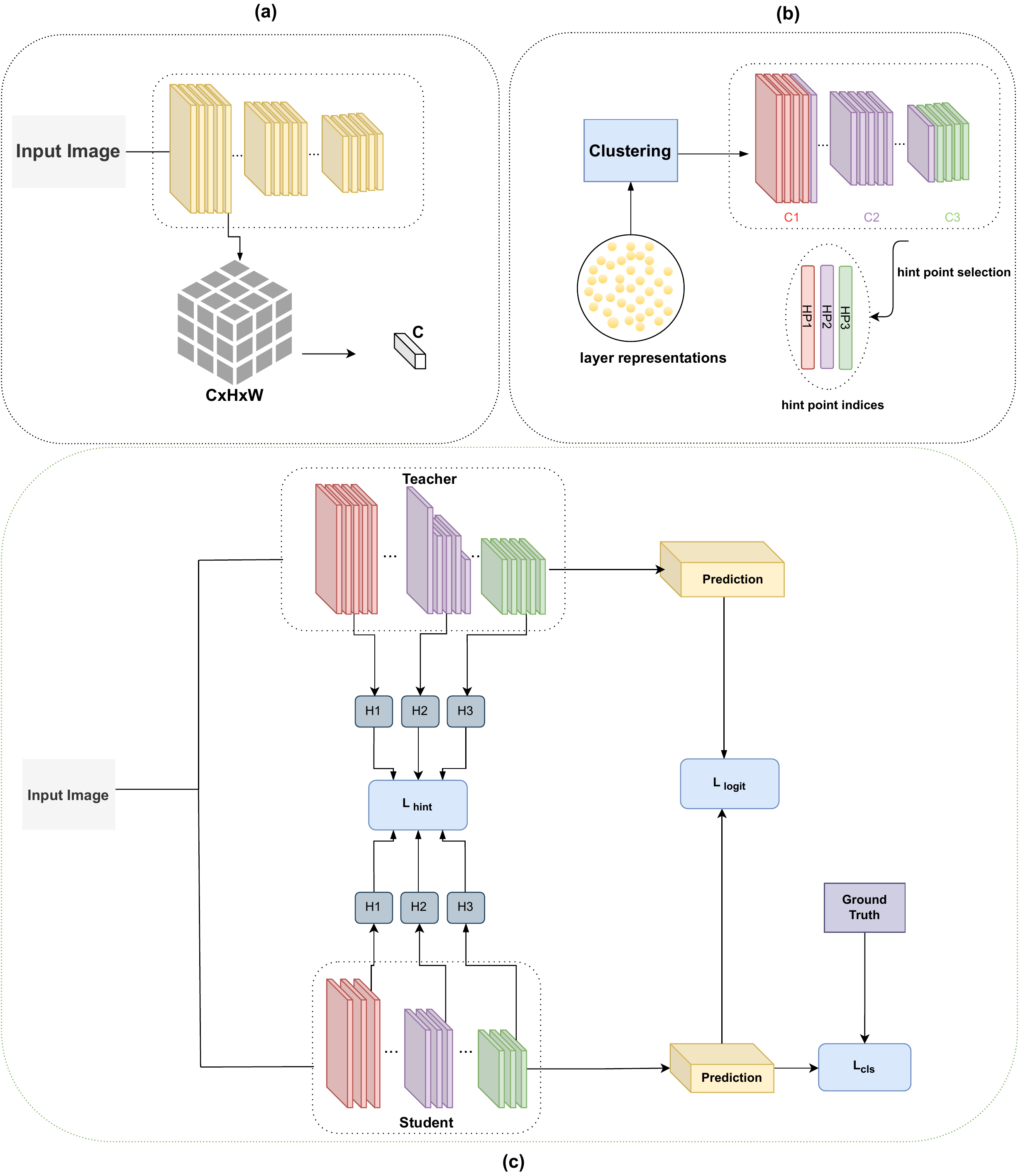}
		\caption{The workflow of PURSUhInT consists of three steps: (a) 
		The acquisition of layer representations. For this purpose, feature maps for a certain number of samples ($N$) are obtained from all of the sub-blocks of pre-trained teacher and averaged in dimensions of height and width to form matrices with the size of $N \times C$. 
		(b) Clustering layer representations to obtain clusters of sub-blocks utilizing k-means. To remove redundancy, we choose one hint point for each cluster, specifically center positions of the clusters. 
		(c) Distillation with the determined hint positions in the previous step. In the diagrams, HP represents the index of hint point, H depicts the transformation which determines the hint type and CL shows the determined cluster. Best viewed in color.}
		\label{fig:all_steps}
	\end{figure}

In this paper, we demonstrate hint point selection's effect on the performance of knowledge distillation and propose PURSUhInT, an informative hint position selection method based on layer clustering. To that end, we employ pre-trained teacher models and apply clustering on their sub-blocks. For this purpose, we first obtain the layer representations of each sub-block. Then we utilize k-means~\citep{k-means}~clustering algorithm on these representation matrices to obtain clusters of sub-blocks. This process yields clusters which consist of similar sub-blocks in terms of the predefined metric. Consequently, we choose one hint position per cluster to distill teacher network's knowledge in a comprehensively accurate fashion. Figure~\ref{fig:all_steps} demonstrates steps of our method. In this section, we elaborate on these steps.

\subsection{Layer Representation}
In order to cluster the sub-blocks of networks, sub-blocks should be represented in a proper form. In this study, we follow the commonly used setup for representing layers~\citep{kornblith2019similarity, raghu2017svcca,morcos2018insights}, which will be described in this section.

Let $H_i$ be the feature map of $i^{th}$ sub-block in a teacher network which consists of blocks such as ResNet. Its size is $N \times C \times H \times W$, where $N$ is the number of samples, $C$ is the number of channels, $H$ and $W$ are the height and width of the feature map, respectively. Since spatial dimensions share parameters, dimension of the channel contains more information. Hence, feature maps are averaged in dimensions of height and width to obtain representation matrices with the size of $N \times C$. 
To apply clustering, we store feature maps of the sub-blocks for $10^4$ samples from the training set. 

\subsection{Clustering of Layers}

The main approach adopted in the proposed knowledge distillation method is layer clustering. Clustering algorithms necessitate specification of a similarity metric defined among different clusters. Similarities between different layers of an artificial neural network are determined by methods exploiting various attributes of the network, such as, weight matrices~\citep{neill2020compressing}, activation maps~\citep{kornblith2019similarity, raghu2017svcca}, and the layers with the same order at different networks~\citep{li2015convergent}.~\citep{li2015convergent} proposes using correlation as a similarity metric to compare the knowledge learned by different networks with the same structure. They point out that using correlation metric for comparison yields similar results with mutual information. In~\citep{raghu2017svcca}, authors proposed using canonical correlation analysis (CCA) to measure the similarity between layers.  Moreover,~\citep{morcos2018insights} improved this metric by weighting. Then,~\citep{kornblith2019similarity} proposes centered kernel alignment (CKA) as a similarity index, whose linear case is associated with CCA.

In this paper, we use CKA and mean squared CCA ($R^2_{CCA}$) as the similarity metrics for layers in a neural network, where $R^2_{CCA}$ is a statistic that shows convergence of CCA~\citep{kornblith2019similarity}. Moreover, $R^2_{CCA}$ is acknowledged as Yanai’s GCD measure~\citep{ramsay1984matrix}.

Mean squared CCA can be calculated by~(\ref{eq:R2_CCA}) as follows:
\begin{equation}
\label{eq:R2_CCA}
    R^2_{CCA}(X,Y) = \frac{ \sum_{i=1}^{p_1} \rho^2_i}{p_1} = \frac{{\left \| Q_Y^T Q_X \right \|}_F^2}{p_1}
\end{equation}

\noindent where $\rho_i$ is the $i^{th}$ canonical correlation coefficient, $p_1$ shows the minimum of shapes of the components, $Q_Y$ and $Q_X$ are the orthonormal bases for the columns of $Y$ and $X$, respectively.

Furthermore, CKA is computed by~(\ref{eq:CKA}) as follows:
\begin{equation}
\label{eq:CKA}
    CKA(K,L) = \frac{ HSIC(K,L)}{  \sqrt{HSIC(K,K) HSIC(L,L)}}
\end{equation}
where HSIC is Hilbert-Schmidt independence criteria. The empirical estimator of HSIC is shown in~(\ref{eq:HSIC}) as:

\begin{equation}
\label{eq:HSIC}
    HSIC(K,L) = \frac{1}{(n-1)^2}tr(KHLH)
\end{equation}

\noindent where H is the centering matrix, tr represents trace of the matrix and n is the number of samples. $K_{ij} = k(x_i,x_j)$ and $L_{ij} = l(y_i,y_j)$ where k and l are two kernels. In this paper we use linear kernels unless it is explicitly stated that RBF kernel is used.

In order to distill teacher's knowledge successfully, hints should carry concise information. By utilizing $R^2_{CCA}$ and $CKA$ to group layers of the teacher network, we obtain clusters of similar layers in terms of these metrics. We provide optimum hint points that contain non-redundant information by selecting one hint point per cluster. For this purpose, we utilized  $1 - R^2_{CCA}$ and $1 - CKA$ as the distance metrics for k-means algorithm where k corresponds to number of hints. 

K-means is a clustering method that aims to partition the n observations into k clusters. To find the best centroid for each cluster, k-means repeats assignment and update steps until the assignments are fixed, where the steps are summarized as follows:

1) Assign the data points based on the labels of the current nearest centroids:

\begin{equation}
\label{eq:y_i}
    y_i = \arg\min_j D(x_i, \mu_j),
\end{equation}
where  $\mu_j$ represents the centroid of $j^{th}$ cluster. $x_i$ and $y_i$ are the features and assigned cluster of the $i^{th}$ data point, respectively. D is the chosen distance metric, considering teacher model and the hint distillation type.

2) Determine the centroids based on the current assignment of data points for each cluster:

\begin{equation} 
\label{eq:mu_j} 
    \mu_j=\frac{1}{|C_j|}\sum_{x_i \in C_j} x_i,  
\end{equation}
where $C_j$ shows the $j^{th}$ cluster.  \\ 

{
We choose k-means as our clustering algorithm since it's simple and effective. Although it has disadvantages for other applications, these disadvantages are not valid for our problem, since the number of clusters is predetermined as the number of hints and our data is not high-dimensional. Moreover, to mitigate the seeding problem, we choose three layers farthest from each other considering their order in the model, which are the first, center and last layers (points), as initial seeds.}

{
Table \ref{tab:exec_time} presents the execution times for the first two steps of our workflow, namely, acquisition of layer representations and clustering these representations, for the teacher models used in experiments on CIFAR-100. It should be highlighted that running the first two steps once is sufficient. We observe that clustering with the similarity metric of $R^2_{CCA}$ takes a shorter time than that of $CKA$.
}

\begin{table}[h]
\centering
\caption{{Execution times for Step 1 and Step 2 in our workflow, where results are obtained on CIFAR-100 dataset. Step 2 is achieved with k-means which uses the stated similarity metrics. It should be noted that the execution times for the clustering step are acquired by using only CPU, where this step can be accelerated by utilizing GPU.
\\}}
\label{tab:exec_time}

\begin{tabular}{c|c|c|c|c}

{Teacher} & {Number of } & {Similarity} & \multicolumn{2}{c}{Execution time (s)} \\ \cline{4-5}
model & layers & metric & Step 1 & Step 2 \\ \hline\hline

WRN-40-2 & 18 & $R^2_{CCA}$ & 272 & 180 \\ \hline
resnet32x4 & 15 & $R^2_{CCA}$ & 239 & 220 \\ \hline

\multirow{2}{*}{resnet110} & \multirow{2}{*}{54} & $R^2_{CCA}$ & \multirow{2}{*}{941} & 336 \\
&&$CKA$ & & 4797 \\
\hline
\end{tabular}
\end{table}

\subsection{Hint Distillation with Hint Point Search}

%Hint distillation methods are used for improving logit distillation. Loss function used in these studies can be defined as in~(\ref{eq:loss}):

Hint distillation methods are mostly used for improving logit distillation, where student model receives more guidance from teacher with the knowledge from intermediate layers, namely, hints. Hence, loss function consists of multiple terms that guide several layers of student model \citep{peng2019correlation, yang2022multi}. Therefore, total loss function, $L_{total}$, used for training the student model can be defined as follows:

%\begin{equation}
%\label{eq:loss}
%    Loss = \gamma L_{cls} + \alpha L_{logit} + \beta L_{hint}
%\end{equation}
{
\begin{equation}
\label{eq:loss}
    L_{total} = \gamma L_{cls}(p^S,y) + \alpha L_{logit}(p^S,p^T) + \beta L_{hint}(f^S,f^T)
\end{equation}
}where $L_{logit}$ and $L_{cls}$ represent the output (logit) distillation loss and classification loss, respectively. In~(\ref{eq:loss})
$y$, $p$ and $f$ denote labels, logits and hints obtained from models, respectively. Furthermore, $T$ and $S$ superscripts represent teacher and student models, respectively.
Moreover, $\gamma$, $\alpha$ and $\beta$ are weights for losses of classification, logit distillation and hint distillation, respectively. 
%We define hint distillation loss \textcolor{red}{$L_{hint}$} in~(\ref{eq:l_hint}) as the following:  

Hint distillation loss {$L_{hint}$} can be formulated as in the following:
\begin{equation}
\label{eq:l_hint}
    L_{hint} = \sum_{i}^k L(F(S_i),F(T_i)),  
\end{equation}
where $k$ is the number of hints, $L$ is the predetermined loss function for hint transfer, $F$ represents the transformation which determines the hint type, $T_i$ and $S_i$ show features from the $i^{th}$ group of layers in teacher and student networks, respectively. 
It should be noted that, as opposed to most of the previous studies operating on the transformation $F$, our method alters and re-defines the grouping approach.

Studies done so far uses groups of layers with the same spatial size. However, in our experiments we observe that some of these conventional hint positions fall into the same cluster that are determined by specific metrics designed for layer similarity (See Figure \ref{fig:hint_pos}). Our experiments demonstrate that selection of hint positions plays an important role on distillation performance. Hence, we propose to determine groups of layers by clustering using proper metrics in order to obtain effective hint positions. Our method prevents using redundant information obtained from similar layers for distillation.

Moreover, we consider using the center points to represent the groups of layers instead of the last points as in conventional approach. Hence, we conducted experiments for these points with different metrics as presented in Table~\ref{tab:AT_all_metrics}. {The first points in clusters are not included in these experiments, since choosing the first positions/layers in clusters yields transferring knowledge from the first layers of networks for the first hint points. This would result in transferring of redundant knowledge since features on the first layers of the model are general and mostly unrelated to the objective function \citep{yosinski2014transferable}.}

Our experiments show that center points yield better results than the last points of clusters which we obtain. Furthermore, the results point out that changing hint points to center without the proposed grouping, would yield decrease of performance instead of improvement.

\begin{table}[h]
\centering
\caption{Top-1 accuracy (\%) results of distilled resnet20 for different hint positions in baseline grouping and clusters obtained with different metrics, on CIFAR-100. Attention Transfer is used  as the distillation type in these experiments. * stands for the normalization before clustering. Although $R^2_{CCA}$ yields different clusters for with and without normalization, the center points are the same for these cases.\\}
\label{tab:AT_all_metrics}

\begin{tabular}{l|c|c}

{Metric type} & {Position on cluster} & {Accuracy}     \\ \hline\hline
\multirow{2}{*}{- (baseline)} & last & 71.35 ± 0.43 \\ %same spatial size
& center & 71.29 ± 0.15 \\ \hline

\multirow{2}{*}{$R^2_{CCA}$*} & last & 71.22 ± 0.13 \\ 
 & center & 71.59 ± 0.32 \\ \hline
 
\multirow{2}{*}{$R^2_{CCA}$}  & last & 70.99 ± 0.16 \\ 
 & center & 71.59 ± 0.32\\ \hline
 
\multirow{2}{*}{$CKA_{linear}$ or $CKA_{rbf}$}& last & 71.20 ± 0.17 \\ 
 & center & 71.50 ± 0.14\\ \hline

\end{tabular}
\end{table}

\section{Experiments}
\label{sec:experiments} 

We conduct experiments on CIFAR-100 and ImageNet, which are commonly used datasets in knowledge distillation studies. Moreover, we utilize a wide variety of distillation methods as well as architectures for teacher and student models such as ResNet~\citep{resnet}, Wide ResNet~\citep{zagoruyko2016wide} and ShuffleNet~\citep{zhang2018shufflenet}. Besides, we evaluate our approach on both of pre-activation and post-activation hints.

\begin{figure}
\centering

\begin{subfigure}{\linewidth}
    \centering
    \includegraphics[height=1.4in]{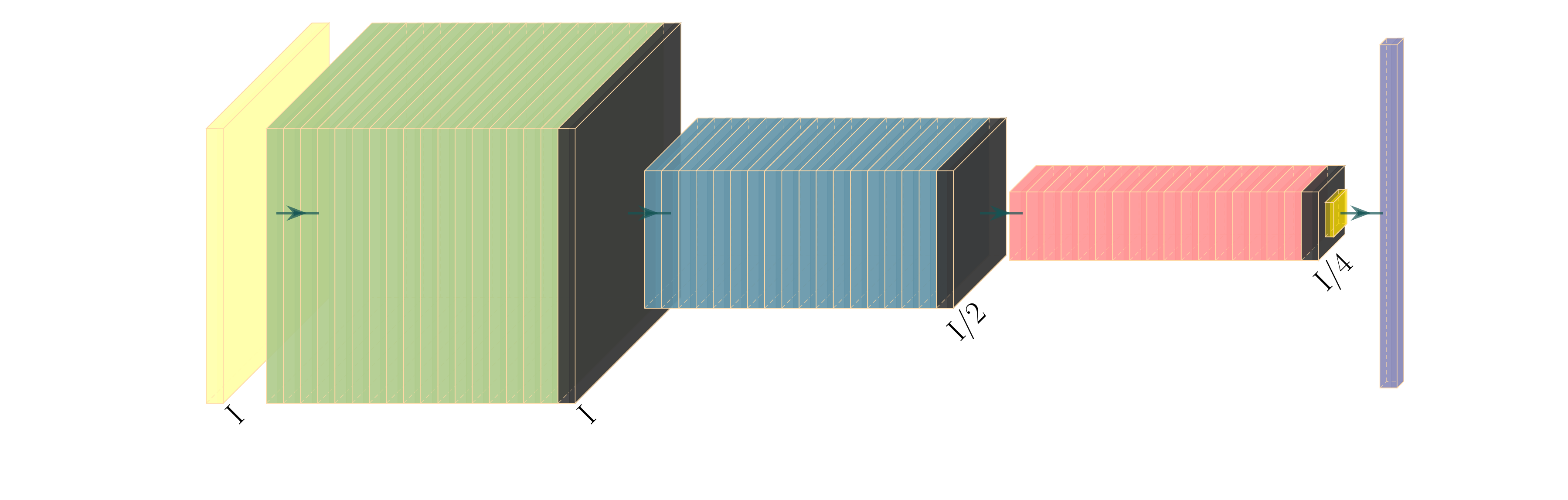}
    \caption{}
   \end{subfigure}

\hfill

  \begin{subfigure}{\linewidth}
  \centering
  \includegraphics[height=1.4in]{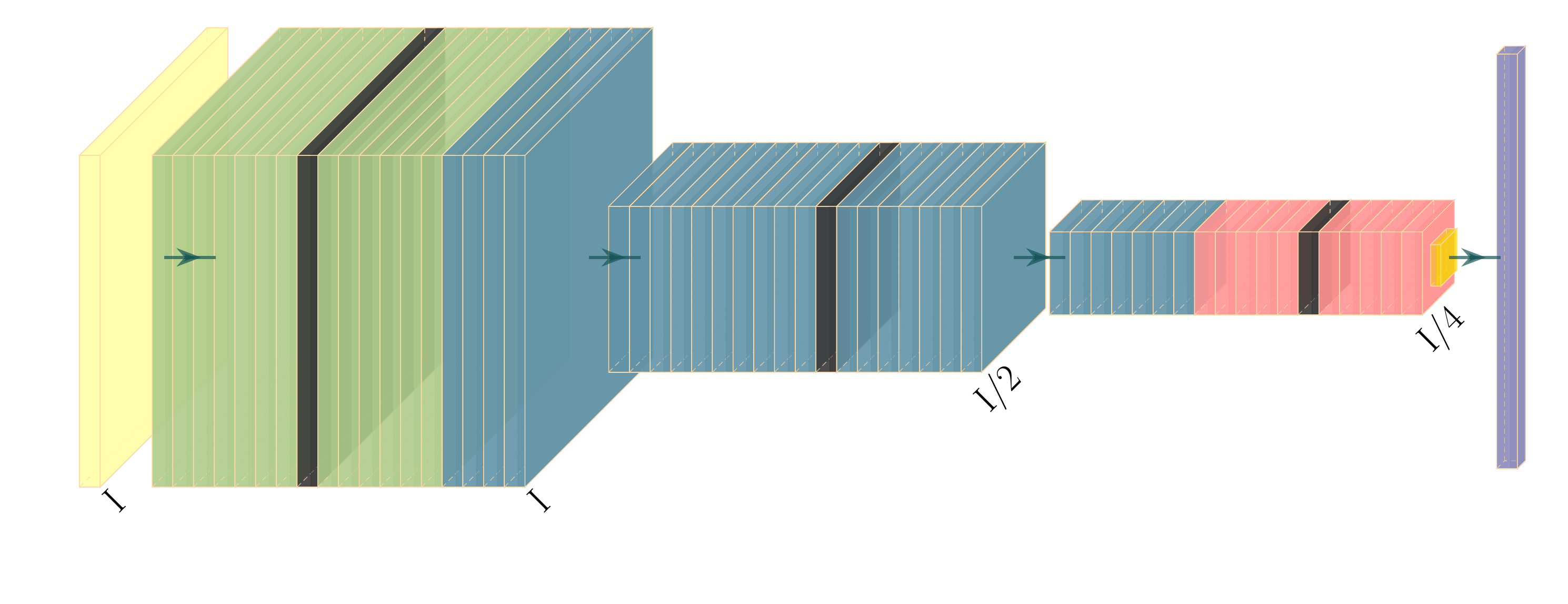}
  \caption{}
  \end{subfigure} 

\RawCaption{\caption{Representative structure of ResNet models with hint positions which are colored as black. The labels at the bottom right of the blocks show the spatial sizes of feature maps, where $I$ corresponds to the spatial size of the input image. 
(a) Common approach for grouping layers of ResNets which only considers spatial size of the feature maps. The last points of these groups are chosen as the hint points. (b) Our proposed approach which utilizes clustering with the aim of focusing on layer similarity. This figure depicts the clusters obtained by our method with ${CKA}$ metric on resnet110 sub-blocks, for CIFAR-100 dataset. We choose the center points of each cluster as hint positions for distillation of teacher networks. It is visualized using \citep{harisIqbal} and best viewed in color.}
\label{fig:hint_pos}}
\end{figure}

As stated in Section \ref{sec:method}, our method consists of three steps. After obtaining layer representations of candidate hint positions, we cluster these representations using k-means algorithm. We choose $k=3$ for experiments on CIFAR-100 and $k=4$ for experiments on ImageNet, in order to be consistent with the conventional number of hints. After the clustering, we choose center points of the clusters as proposed hint positions. Table \ref{tab:HP} highlights the difference between conventional hint positions and proposed hint positions by our approach. Besides, performances of the teacher models used in our experiments are presented. It is observed that some of the conventional hint positions may take place in the same clusters obtained by our method as shown in Figure \ref{fig:hint_pos}. This yields transferring redundant and non-efficient information from teacher to student models in conventional approaches. The last step in our approach is distillation with the proposed hint positions, which is explained in detail in Section \ref{subs:exp_set}.

\begin{table}[h]
\centering
\caption{Comparison of baseline and proposed hint positions (HPs) for teacher models used in our experiments. Proposed HPs are obtained by our algorithm which utilizes k-means with the specified metrics as the distance functions. Moreover, the datasets used for the testing of the teacher models and the accuracies obtained by the teacher models are presented. AT presents Attention Transfer \citep{Zagoruyko2017AT} as the hint distillation method. 
\\}
\centerline{
\begin{tabular}{l|c|c|c|c}
Teacher Model & Dataset & Accuracy & Baseline HPs & Proposed HPs % & Metric 
\\ \hline

Pre-activation hints: & & & \\ \hline
resnet110 (for AT) & CIFAR-100 & 74.49 & 18, 36, 54 & 6,27,49 \\% & $R^{2}_{CCA}$ \\ %R2
resnet110 (for others) & CIFAR-100 & 74.49 & 18, 36, 54 & 8,29,49 \\% & $CKA$ \\ %CKA &(VID, FitNets) 
\hline
Post-activation hints: & & & \\ \hline
resnet110  & CIFAR-100 & 74.31 & 18, 36, 54    & 7, 29, 49 \\% & $CKA$      \\ %CKA_lin
resnet32x4 & CIFAR-100 & 79.42 & 5, 10, 15     & 3, 8, 13  \\% & $R^{2}_{CCA}$      \\ %R2
WRN-40-2   & CIFAR-100 & 75.61 & 6, 12, 18     & 1, 8, 16  \\% & $R^{2}_{CCA}$      \\ %R2
ResNet-34  & ImageNet & 73.31 & 3, 7, 13, 16   & 2, 6, 11, 15 \\% & $CKA$ \\  %CKA_lin
\hline

\end{tabular}}
\label{tab:HP}
\end{table}

%sonuçlar:
% -imagenet
% -cifar
    % -preact (rn110->rn8/20/32)
    % -postact (SOTA + CRD tablosu)

\subsection{Experimental Settings}
\label{subs:exp_set}

In the experiments, the similarity metrics for clustering of layers are determined considering distillation performances. For the implementation of hint types, we follow the setup of \citep{tian2019contrastive}, except FitNets \citep{romero2015fitnets}. In this study, we use three hints for FitNets for fair comparison with other hint types. Moreover, we determine weights of loss components by Bayesian search~\citep{wandb}, for all methods. Additionally, we choose the query-key dimension for AFD method~\citep{ji2021show}.

For experiments on pre-activation hints, we use pre-ReLU hints for all methods and SGD as the optimizer. Moreover, we set the maximum iteration as 350 epochs and batch size as 64. Besides, initial learning rate is 0.05 and it is decayed by 0.1 every 50 epochs after the first 200 epochs. For experiments on post-activation hints, we use post-ReLU hints for all methods and SGD as the optimizer. Furthermore, we set the maximum iteration as 240 epochs for fair comparison with other approaches and determine other hyperparameters by Bayesian search, that are temperature, learning rate and batch size.

For ImageNet experiments, we use Attention Transfer as the hint type with $p=1$ \citep{Zagoruyko2017AT}. We set the maximum iteration as 100 epochs, batch size as 256 and temperature as 4, by following \citep{tian2019contrastive}. We use SGD as the optimizer. Besides, initial learning rate is 0.1 and it is decayed by 0.1 at epochs of 30, 60, 80 and 90.

\subsection{Results on CIFAR-100}

To evaluate our approach on CIFAR-100 dataset, we conduct experiments on pre-activation and post-activation hints, which are the hints obtained before and after the last ReLU function in a sub-block, respectively. 

\subsubsection{Pre-activation hints}

For experiments on pre-activation hints, we choose three hint distillation methods to compare our findings, which are FitNets \citep{romero2015fitnets}, Attention Transfer (AT) \citep{Zagoruyko2017AT} and Variational Information Distillation (VID) \citep{ahn2019variational}. We choose two logit distillation methods to use along hint distillation methods, which are Hinton's Knowledge Distillation (KD) \citep{hinton2015distilling} and Spherical Knowledge Distillation (SKD) \citep{guo2020spherical}. Thus, we set up 6 experiments for each teacher-student pair. To assess our performance, we compare our proposed hint positions with the baseline hint positions, which are the last layers before the downsampling. We use CIFAR-style ResNets as the teacher and student models for distillation. Moreover, we utilize $R^2_{CCA}$ metric for AT (Attention Transfer) type of distillation and $CKA$ metric for two other types of hint distillation, in these experiments. Baseline and proposed hint positions are presented in Table~\ref{tab:HP}.

\begin{figure}
     \centering
     \begin{subfigure}[b]{0.65\textwidth}
         \centering
         \includegraphics[width=\textwidth]{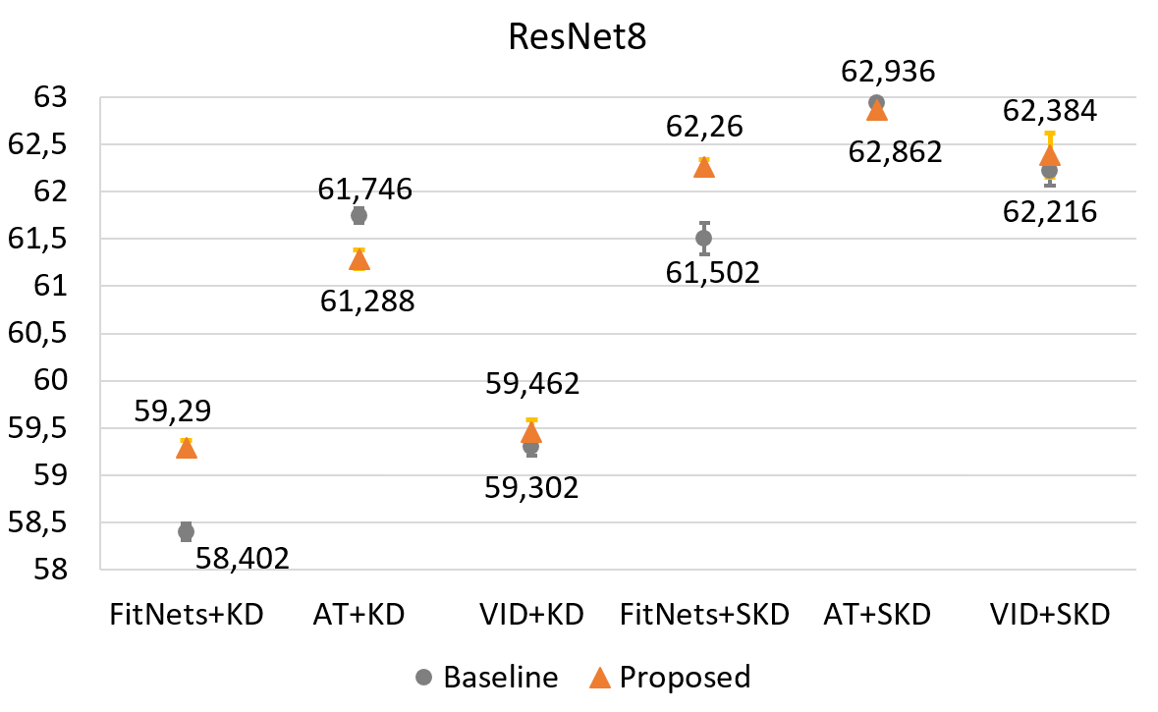}
         \caption{}
         \label{fig:RN8}
     \end{subfigure}
     \hfill
     \begin{subfigure}[b]{0.65\textwidth}
         \centering
         \includegraphics[width=\textwidth]{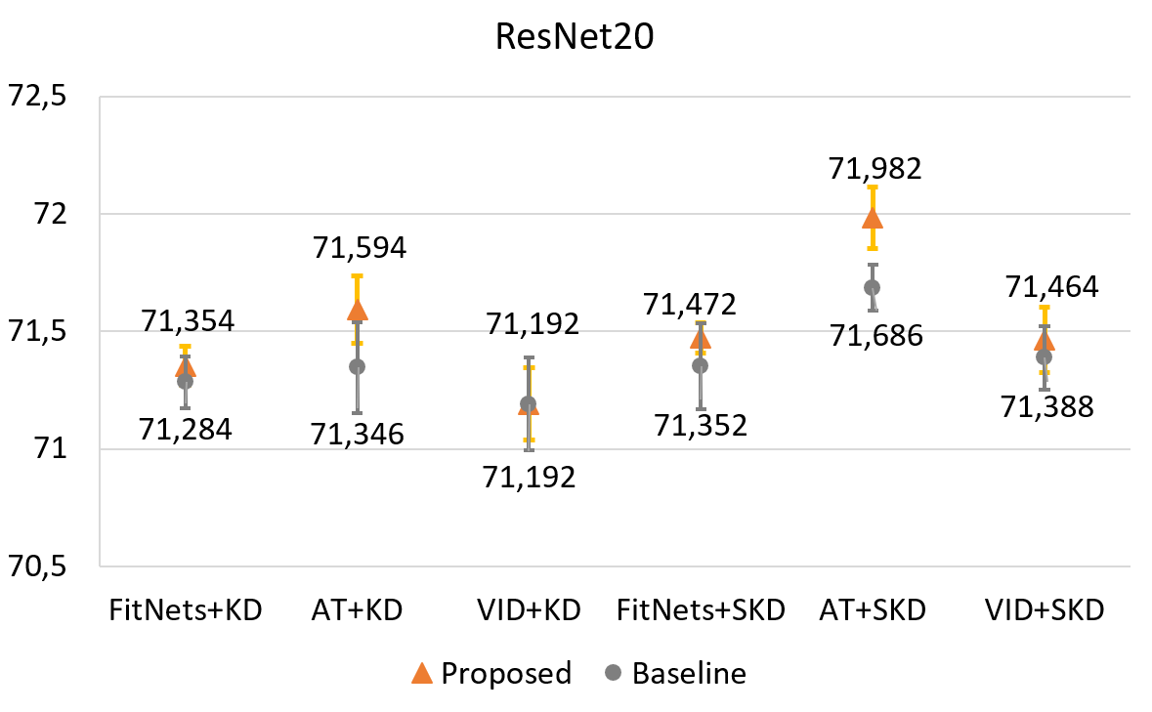}
         \caption{}
         \label{fig:RN20}
     \end{subfigure}
     \hfill
     \begin{subfigure}[b]{0.65\textwidth}
         \centering
         \includegraphics[width=\textwidth]{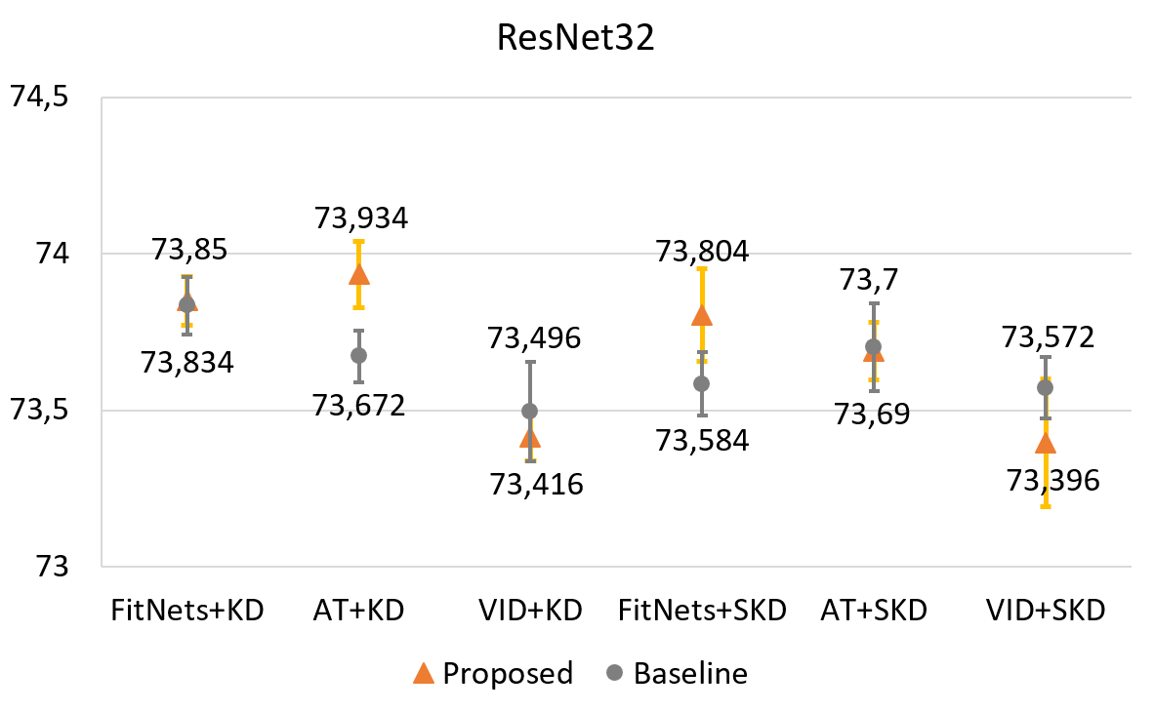}
         \caption{}
         \label{fig:RN32}
     \end{subfigure}
        \caption{{
        Top-1 accuracy (\%) results of (a) resnet8, (b) resnet20 and (c) resnet32 student models for 6 distillation schemes. Teacher model's accuracy is 74.490. Each score is obtained over 5 runs. (Best viewed in color.) %Values in bold show the best accuracies for each distillation scheme.
        \\}}
        \label{fig:RN8-20-32}
\end{figure}

Since our method provides particular hint positions for the teacher network, proposed hint positions are valid for any student model with the same number of hints. Hence, using the same hint positions we conduct experiments for three student models, which are resnet8, resnet20 and resnet32. {Figure \ref{fig:RN8-20-32}} presents the results of knowledge distillation from resnet110 to three student models, for our proposed hint points and the baseline hint points.

Results show that our method outperforms the conventional approach for all hint types on resnet20. Moreover, it is demonstrated that our method yields the best result for distillation of resnet32. 
Besides, it outperforms the baseline except AT type of distillation of resnet8, while it achieves very close performance to the best result. In addition, our method significantly improves the performance of FitNets for distillation of resnet8. 

\subsubsection{Post-activation hints}

For experiments on post-activation hints, we utilize six teacher-student pairs. We apply our method on Attention Transfer (AT) along with the conventional KD, which is described as AT+KD, where we utilize baseline and proposed hint positions for AT hints. Moreover, we employ our approach on the state-of-the-art distillation approaches to boost their performance. For this purpose, we choose two recently proposed methods, namely, Weighted Soft Label (WSL) \citep{zhou2021rethinking} and Attention-based Feature Distillation (AFD) \citep{ji2021show}. WSL is a recent logit distillation method that applies sample-wise weighting on the loss function. To conduct experiments on hint position, we first improve this method by combining it with AT. 
To combine these methods, we simply utilize the loss function in (\ref{eq:loss}) where $L_{hint}$ indicates the loss for AT hints and other terms correspond to the loss of WSL method.
Then, we obtain the results of WSL+AT using baseline and proposed hint positions for transferring AT hints.
Another method that we employ for the experiments is AFD, which utilizes attention mechanism to search efficient hint positions between teacher and student models. It transfers knowledge between these determined positions as well as the logits and repeats the search process during training. This method assumes all intermediate output positions as potential hint positions for establishing connections between models. While it searches efficient points among the all intermediate output positions, we apply our approach on this method by constraining its potential hint positions of teacher as our proposed three hint positions. 

The results on state-of-the-art methods are presented in Table~\ref{tab:AFD_WSL}, while the performances of the teacher models used in the experiments are listed in Table~\ref{tab:HP}. We utilize $CKA$ metric for resnet110 and $R^2_{CCA}$ metric for resnet32x4 and WRN-40-2 teacher models, to obtain proposed hint positions. Results show that the proposed hint position method outperforms the baseline in most cases. Moreover, it yields an increase in accuracy up to 1.5 \% compared to the baseline hint positions. Besides, it consistently improves distillation performance for resnet32x4, resnet20 and resnet8x4 models where first one is a teacher and others are student models. Although we limit the search space of AFD by constraining it as three proposed hint positions, proposed method mostly yields better distillation performance. Specifically, proposed approach improves WSL+AT method except resnet32 student model and AFD method except one teacher-student pair which is WRN-40-2 - ShuffleNetV1. Furthermore, it yields 6.482 \% increase in accuracy for ShuffleNetV1 student, compared to the vanilla training.

\begin{table}[]
\begin{center}

\caption{
Top-1 accuracy (\%) results of student models using state-of-the-art distillation methods on CIFAR-100 dataset. The results are compared between proposed and baseline hint positions for each distillation method. Values in bold show the best accuracies for each distillation scheme. Each score is obtained over 5 runs. Column marked by * show the reported results in \citep{shao2021multi}.
\\}

\begin{tabular}{c|c|c|c|c}
 &  &    & \multicolumn{2}{c}{AT + KD}               \\ \cline{4-5}
 Teacher &   Student &  Vanilla & Baseline  & Proposed    \\    \hline \hline    
\multirow{2}{*}{resnet110}  & resnet20   & 69.06    & 70.95 & \textbf{71.06}     \\
 & resnet32 & 71.14  & \textbf{73.59}     & 73.48              \\ \hline
\multirow{2}{*}{resnet32x4} & ShuffleNetV1 & 70.50   & 75.49 & \textbf{76.98}     \\
  & resnet8x4   & 72.50  & 75.06     & \textbf{76.07}     \\ \hline
\multirow{2}{*}{WRN-40-2}   & WRN-16-2   & 73.26       & \textbf{75.64}     & 75.45         \\
   & ShuffleNetV1  & 70.50    & 76.27              & \textbf{76.78}     \\ \hline \hline 
\multicolumn{1}{l}{}        & \multicolumn{1}{l}{} & &\multicolumn{2}{c}{WSL + AT}    \\ \cline{4-5}
\multicolumn{1}{l}{}        & \multicolumn{1}{l}{} & {WSL}    & Baseline & Proposed \\ \hline
\multirow{2}{*}{resnet110}        & resnet20             & {71.24} & 71.45              & \textbf{71.63}     \\
 & resnet32  & {73.49} & \textbf{73.87}     & 73.73 \\ \hline
\multirow{2}{*}{resnet32x4} & ShuffleNetV1         & {74.33} & 75.95              & \textbf{76.98}     \\
                            & resnet8x4            & {75.24} & 75.04              & \textbf{75.69}     \\ \hline
\multirow{2}{*}{WRN-40-2}   & WRN-16-2             & {75.56} & 75.49              & \textbf{75.60}     \\
                            & ShuffleNetV1            & {75.29} & 75.84              & \textbf{76.52}     \\ \hline \hline 
\multicolumn{1}{l}{}        & \multicolumn{1}{l}{} &    & \multicolumn{2}{c}{AFD}    \\ \cline{4-5}
\multicolumn{1}{l}{}        & \multicolumn{1}{l}{} && {Baseline*}                    & Proposed            \\ \hline
\multirow{2}{*}{resnet110}  & resnet20             & &{71.20}                       & \textbf{71.49}     \\
& resnet32          &   & {73.46}               & \textbf{74.05}     \\ \hline
\multirow{2}{*}{resnet32x4} & ShuffleNetV1         & &{75.08}                       & \textbf{75.11}     \\
& resnet8x4   & &{74.72}        & \textbf{75.80}     \\ \hline
\multirow{2}{*}{WRN-40-2}   & WRN-16-2             & &{75.41}                       & \textbf{75.60}     \\
& ShuffleNetV1  & &{\textbf{75.63}}              & 75.11             \\ \hline
\end{tabular}
\label{tab:AFD_WSL}
\end{center}
\end{table}

For further evaluation, we compare our method with various distillation approaches as~\citep{tian2019contrastive}. For this purpose, we obtain results of our method by integrating it with four distillation approaches, using the same pre-trained teachers as other approaches. Table \ref{tab:CRD} shows the results obtained on the last epochs for fair comparison with \citep{tian2019contrastive}.

Results demonstrate that our approach yields superior performance compared to the various approaches, except one teacher-student pair. It can be observed that the best performances are obtained mostly by AT+KD and CRD+AT+KD approaches with the proposed hint positions. Furthermore, adding CRD transfer to AT+KD yields decrease in performance for resnet32x4 teacher, while it boosts distillation performances of the two teacher-student pairs which are resnet110 - resnet20 and WRN-40-2 - WRN-16-2. Although AFD method with the proposed hint positions is not successful as others for ShuffleNetV1 student, it yields the best result for resnet110 - resnet32 pair. It should be highlighted that our approach on AT+KD yields 1.6 \% improvement over the second best method for teacher-student pair of resnet32x4 - ShuffleNetV1.

\begin{table}[]
\setlength{\tabcolsep}{3.5pt}
\caption{
Top-1 accuracy (\%) results of models using various distillation methods on CIFAR-100 dataset. Each score is obtained over 5 runs. Values in bold show the best accuracies for each distillation scheme.
It should be noted that reported results are based on the last epoch for fair comparison with \citep{tian2019contrastive} where the results of other methods are quoted from. Moreover, some results for FSP~\citep{yim2017gift} method cannot be obtained since it is not applicable for student-teacher pairs with different architectures. 
\\}
\centerline{
\begin{tabular}{l|c|c|c|c|c|c}
Teacher      & {WRN-40-2} & {resnet110} & {resnet110} & {resnet32x4} & {resnet32x4} & {WRN-40-2} \\
Student    & WRN-16-2    & {resnet20}  & {resnet32}  & resnet8x4& ShuffleNetV1  & ShuffleNetV1 \\ \hline
\multirow{2}{*}{Vanilla}    & 75.61  & 74.31   & 74.31   & 79.42    & 79.42    & 75.61  \\
    & 73.26  & 69.06   & 71.14   & 72.50    & 70.50     & 70.50   \\ \hline
KD  \citep{hinton2015distilling}   & 74.92  & 70.67   & 73.08   & 73.33    & 74.07    & 74.83  \\
FitNet \citep{romero2015fitnets} & 73.58  & 68.99   & 71.06   & 73.50    & 73.59    & 73.73  \\
AT \citep{Zagoruyko2017AT}         & 74.08  & 70.22   & 72.31   & 73.44    & 71.73    & 73.32  \\
SP \citep{tung2019similarity}        & 73.83  & 70.04   & 72.69    & 72.94    & 73.48      & 74.52 \\
CC \citep{peng2019correlation}        & 73.56  & 69.48  & 71.48    & 72.97    & 71.14        & 71.38   \\
VID \citep{ahn2019variational}       & 74.11  & 70.16   & 72.61     & 73.09    & 73.38  & 73.61  \\
RKD \citep{park2019relational}       & 73.35  & 69.25   & 71.82    & 71.90    & 72.28    & 72.21  \\
PKT \citep{passalis2018learning}       & 74.54  & 70.25   & 72.61       & 73.64    & 74.10    & 73.89   \\
AB \citep{heo2019knowledge}        & 72.50    & 69.53   & 70.98   & 73.17    & 73.55    & 73.34  \\
FT \citep{kim2018paraphrasing}    & 73.25  & 70.22   & 72.37   & 72.86    & 71.75    & 72.03  \\
FSP  \citep{yim2017gift}  & 72.91  & 70.11   & 71.89   & 72.62    & n/a & n/a \\
NST  \citep{huang2017like}  & 73.68  & 69.53   & 71.96   & 73.30    & 74.12    & 74.89  \\
CRD \citep{tian2019contrastive}       & {{75.48}}   & {\textbf{71.46}}    & {73.48}    & {75.51}     & {75.11}     & {76.05}   \\

\hline
%last:
Ours (AT+KD) & 75.16 & 70.86 & 73.12 & \textbf{75.84}  & \textbf{76.74} & {76.55} \\
Ours (WSL+AT) & 75.39 & 71.43 & 73.44  & 75.49 & 76.70 & 76.36 \\
Ours (AFD) & 75.33 & 71.22 & \textbf{73.79} & 75.52 & 74.94  & 74.94 \\  
Ours (CRD+AT+KD) & {\textbf{75.82}} & {71.32} & {73.19} & {74.20} & {74.88} & {\textbf{76.66}} \\
% &&&&&&\\
\hline
%metin içine de çekilebilir bu atıflar
\end{tabular}}
\label{tab:CRD}
\end{table}

\subsection{Results on ImageNet}
To further assess the performance of our approach, we use ImageNet which is one of the large-scale datasets. For this purpose, Attention Transfer (AT) is utilized as the hint type to be transferred, and KD is used for logit distillation, where the experiment is described as AT+KD. For this experiment, ImageNet-style ResNet-34 and ResNet-18 architectures are used as teacher and student models as in \citep{tian2019contrastive}, respectively. We utilize $CKA$ metric for clustering on ResNet-34 teacher model in order to obtain the proposed hint positions {for AT hints}. Table \ref{tab:imagenet} shows the results compared with the other knowledge distillation methods, where the results of other methods are quoted from \citep{tian2019contrastive}. Moreover, determined hint positions by our method are listed in Table \ref{tab:HP}.

\begin{table}[t]
\caption{Top-1 and Top-5 error rates on ImageNet validation set. ResNet-34 and ResNet-18 architectures are used as teacher and student models, respectively. Results of other approaches are quoted from \citep{tian2019contrastive}, { where other approaches consist of AT \citep{Zagoruyko2017AT}, KD \citep{hinton2015distilling}, SP \citep{tung2019similarity}, CC \citep{peng2019correlation}, Online KD \citep{lan2018knowledge} and CRD \citep{tian2019contrastive}.} \\
}
\setlength{\tabcolsep}{3.5pt}

\centerline{
\begin{tabular}{l|c|c|c|c|c|c|c|c|c|c}
 
 & \multirow{2}{*}{
 Teacher} & \multirow{2}{*}{Student} & \multirow{2}{*}{AT} & \multirow{2}{*}{KD} & \multirow{2}{*}{SP} & \multirow{2}{*}{CC} & Online  & \multirow{2}{*}{CRD} & \multirow{2}{*}{CRD+KD} & Ours \\
 &&&&&&& KD &&&(AT+KD) \\ \hline
 
Top-1 & 26.69 & 30.25 & 29.30 & 29.34 & 29.38 & 30.04 & 29.45 & 28.83 & {28.62} & \textbf{28.51} \\ \hline
Top-5 & 8.58 & 10.93 & 10.00 & 10.12 & 10.20 & 10.83 & 10.41 & 9.87 & \textbf{{9.51}} & 9.83 \\ \hline
%best
\end{tabular}}
\label{tab:imagenet}
\end{table}

Results show that our hint position method yields the best Top-1 accuracy among the compared distillation approaches on ImageNet which is a challenging dataset in computer vision field. More importantly, it can be seen that only changing hint positions for a typical hint type may yield outperforming results compared to the recent methods that use contrastive learning.

As shown in the results, our proposed method improves the hint distillation methods which play a significant role in knowledge distillation for model compression. Table~\ref{tab:compare_models} presents the gains of our method in terms of accuracy among inference time and model size. Results demonstrate that resnet 110 inference time can be improved by a factor of 2.99 with only a Top-1 accuracy loss of 0.26\%. Furthermore, resnet32x4 can be compressed into resnet8x4 with the memory gain of 83.4 \% and a speed-up in inference time of 2.34. Moreover, WRN40-2 can be compressed into ShuffleNetV1 with the memory gain of 57.9 \% and an increase in accuracy of 1.27\%. Besides, WRN40-2 can be compressed into WRN-16-2 with the memory gain of 68.8 \% and an increase in accuracy of 0.4\%.
Results on ImageNet show that ResNet-34 can be compressed into ResNet-18 with a speed-up in inference time of 1.55 and memory gain of 46.4 \%.

\begin{table}[h]
\centering
\caption{Number of parameters, inference time per sample and Top-1 accuracy(\%) results for teacher and student networks. Student networks are compared with teacher networks in the aspects of compression ratio and achieved speed-up, where student models are trained using the proposed hint positions.
It should be noted that accuracies of ResNet-34 and ResNet-18 models are obtained on ImageNet.
\\}
\label{tab:compare_models}

\centerline{
\begin{tabular}{l|c|c|c|c|c}

\textbf{\multirow{2}{*}{Networks}} & \textbf{\multirow{2}{*}{\#Parameters}} & {\textbf{Compression }} & \textbf{{Inference}}    &\textbf{\multirow{2}{*}{Speed-up}}  & \textbf{\multirow{2}{*}{Accuracy}}     \\ 
&&\textbf{ratio}&\textbf{time}&&\\
\hline
Teacher: resnet110 & 1.74 M & - & 24.66 ms & - & 74.31 \\ \hline
resnet20 & 278.32 K & 84.0 \% & 5.84 ms & x4.22 & 71.63 \\
resnet32 & 472.76 K & 72.8 \% & 8.25 ms & x2.99 & 74.05 \\ \hline\hline
Teacher: resnet32x4 & 7.43 M & - & 8.00 ms & - & 79.42 \\\hline
ShuffleNetV1 & 949.26 K & 87.2 \% & 13.37 ms & x0.60 & 76.98 \\
resnet8x4 & 1.23 M & 83.4 \% & 3.42 ms & x2.34 & 76.07 \\\hline\hline
Teacher: WRN-40-2 & 2.26 M & - & 10.07 ms & - & 75.61 \\\hline
WRN-16-2 & 703.28 K & 68.8 \% & 4.71 ms & x2.14 & 76.01 \\
ShuffleNetV1 & 949.26 K & 57.9 \% & 13.37 ms & x0.75 & 76.88  \\\hline\hline
Teacher: ResNet-34 & 21.80 M & - & 3.30 ms & - & 73.31 \\\hline
ResNet-18 & 11.69 M & 46.4 \% & 2.13 ms & x1.55 & 71.49\\
\hline
\end{tabular}%
}
\end{table}

\section{Conclusion}

In this paper, we address the grouping problem on teacher networks to determine the hint positions among the network's sub-blocks. For tackling this problem, we employed k-means algorithm with metrics designed for layer similarity in order to cluster these sub-blocks. Our approach is applicable for any hint distillation scenario which uses at least two hints, in offline distillation scheme. Furthermore, it is valid for any student model, once it is applied on a determined teacher model.

To validate our approach, we apply our method on state-of-the-art distillation methods with the comparison of conventional approach and the proposed approach. Moreover, we present a comprehensive comparison among various distillation approaches and methods that utilize the proposed hint positions. Experimental results suggest that our proposed approach outperforms the state-of-the-art algorithms for numerous architectures on CIFAR-100 and ImageNet datasets. Besides, the proposed method performs successfully in terms of model compression, where it may yield high compression ratio, speed-up in inference time and a gain in accuracy, at the same time. For our subsequent work, we are planning to evaluate our method on different tasks such as object detection on COCO dataset.

\section{Acknowledgements}
This work has been supported by Arcelik ITU R\&D Center, 
The Scientific and Technological Research Council of Turkey (TUBITAK) under the grant number 121E378 and ITU Scientific Research Projects Fund under the grant number MOA-2019-42321.

\bibliographystyle{model5-names}
\bibliography{biblist}

\begin{thebibliography}{70}
\expandafter\ifx\csname natexlab\endcsname\relax\def\natexlab#1{#1}\fi
\providecommand{\url}[1]{\texttt{#1}}
\providecommand{\href}[2]{#2}
\providecommand{\path}[1]{#1}
\providecommand{\DOIprefix}{doi:}
\providecommand{\ArXivprefix}{arXiv:}
\providecommand{\URLprefix}{URL: }
\providecommand{\Pubmedprefix}{pmid:}
\providecommand{\doi}[1]{\href{http://dx.doi.org/#1}{\path{#1}}}
\providecommand{\Pubmed}[1]{\href{pmid:#1}{\path{#1}}}
\providecommand{\bibinfo}[2]{#2}
\ifx\xfnm\relax \def\xfnm[#1]{\unskip,\space#1}\fi
%Type = Inproceedings
\bibitem[{Ahn et~al.(2019)Ahn, Hu, Damianou, Lawrence \&
  Dai}]{ahn2019variational}
\bibinfo{author}{Ahn, S.}, \bibinfo{author}{Hu, S.~X.},
  \bibinfo{author}{Damianou, A.}, \bibinfo{author}{Lawrence, N.~D.}, \&
  \bibinfo{author}{Dai, Z.} (\bibinfo{year}{2019}).
\newblock \bibinfo{title}{Variational information distillation for knowledge
  transfer}.
\newblock In {\it \bibinfo{booktitle}{Proceedings of the IEEE Conference on
  Computer Vision and Pattern Recognition}\/} (pp.
  \bibinfo{pages}{9163--9171}).
%Type = Misc
\bibitem[{Biewald(2020)}]{wandb}
\bibinfo{author}{Biewald, L.} (\bibinfo{year}{2020}).
\newblock \bibinfo{title}{Experiment tracking with weights and biases}.
\newblock \URLprefix \url{https://www.wandb.com/} \bibinfo{note}{software
  available from wandb.com}.
%Type = Inproceedings
\bibitem[{Buciluǎ et~al.(2006)Buciluǎ, Caruana \&
  Niculescu-Mizil}]{bucilua2006model}
\bibinfo{author}{Buciluǎ, C.}, \bibinfo{author}{Caruana, R.}, \&
  \bibinfo{author}{Niculescu-Mizil, A.} (\bibinfo{year}{2006}).
\newblock \bibinfo{title}{Model compression}.
\newblock In {\it \bibinfo{booktitle}{Proceedings of the 12th ACM SIGKDD
  international conference on Knowledge discovery and data mining}\/} (pp.
  \bibinfo{pages}{535--541}).
%Type = Inproceedings
\bibitem[{Chen et~al.(2021)Chen, Liu, Zhao \& Jia}]{chen2021distilling}
\bibinfo{author}{Chen, P.}, \bibinfo{author}{Liu, S.}, \bibinfo{author}{Zhao,
  H.}, \& \bibinfo{author}{Jia, J.} (\bibinfo{year}{2021}).
\newblock \bibinfo{title}{Distilling knowledge via knowledge review}.
\newblock In {\it \bibinfo{booktitle}{Proceedings of the IEEE/CVF Conference on
  Computer Vision and Pattern Recognition}\/} (pp.
  \bibinfo{pages}{5008--5017}).
%Type = Article
\bibitem[{Cheng et~al.(2017)Cheng, Wang, Zhou \& Zhang}]{cheng2017survey}
\bibinfo{author}{Cheng, Y.}, \bibinfo{author}{Wang, D.}, \bibinfo{author}{Zhou,
  P.}, \& \bibinfo{author}{Zhang, T.} (\bibinfo{year}{2017}).
\newblock \bibinfo{title}{A survey of model compression and acceleration for
  deep neural networks}.
\newblock {\it \bibinfo{journal}{arXiv preprint arXiv:1710.09282}\/}, .
%Type = Article
\bibitem[{Choudhary et~al.(2020)Choudhary, Mishra, Goswami \&
  Sarangapani}]{choudhary2020comprehensive}
\bibinfo{author}{Choudhary, T.}, \bibinfo{author}{Mishra, V.},
  \bibinfo{author}{Goswami, A.}, \& \bibinfo{author}{Sarangapani, J.}
  (\bibinfo{year}{2020}).
\newblock \bibinfo{title}{A comprehensive survey on model compression and
  acceleration}.
\newblock {\it \bibinfo{journal}{Artificial Intelligence Review}\/},  (pp.
  \bibinfo{pages}{1--43}).
%Type = Article
\bibitem[{Ciaparrone et~al.(2020)Ciaparrone, S{\'a}nchez, Tabik, Troiano,
  Tagliaferri \& Herrera}]{tracking2020}
\bibinfo{author}{Ciaparrone, G.}, \bibinfo{author}{S{\'a}nchez, F.~L.},
  \bibinfo{author}{Tabik, S.}, \bibinfo{author}{Troiano, L.},
  \bibinfo{author}{Tagliaferri, R.}, \& \bibinfo{author}{Herrera, F.}
  (\bibinfo{year}{2020}).
\newblock \bibinfo{title}{Deep learning in video multi-object tracking: A
  survey}.
\newblock {\it \bibinfo{journal}{Neurocomputing}\/},  {\it
  \bibinfo{volume}{381}\/}, \bibinfo{pages}{61--88}.
%Type = Article
\bibitem[{Deng et~al.(2022)Deng, Sun, Newsam \& Wang}]{deng2022distpro}
\bibinfo{author}{Deng, X.}, \bibinfo{author}{Sun, D.}, \bibinfo{author}{Newsam,
  S.}, \& \bibinfo{author}{Wang, P.} (\bibinfo{year}{2022}).
\newblock \bibinfo{title}{Distpro: Searching a fast knowledge distillation
  process via meta optimization}.
\newblock {\it \bibinfo{journal}{arXiv preprint arXiv:2204.05547}\/}, .
%Type = Article
\bibitem[{Fu et~al.(2021)Fu, Li, Liu \& Yang}]{fu2021interactive}
\bibinfo{author}{Fu, S.}, \bibinfo{author}{Li, Z.}, \bibinfo{author}{Liu, Z.},
  \& \bibinfo{author}{Yang, X.} (\bibinfo{year}{2021}).
\newblock \bibinfo{title}{Interactive knowledge distillation for image
  classification}.
\newblock {\it \bibinfo{journal}{Neurocomputing}\/},  {\it
  \bibinfo{volume}{449}\/}, \bibinfo{pages}{411--421}.
%Type = Article
\bibitem[{Guo et~al.(2020)Guo, Chen, Hu, Zhu, He \& Cai}]{guo2020spherical}
\bibinfo{author}{Guo, J.}, \bibinfo{author}{Chen, M.}, \bibinfo{author}{Hu,
  Y.}, \bibinfo{author}{Zhu, C.}, \bibinfo{author}{He, X.}, \&
  \bibinfo{author}{Cai, D.} (\bibinfo{year}{2020}).
\newblock \bibinfo{title}{Spherical knowledge distillation}.
\newblock {\it \bibinfo{journal}{arXiv preprint arXiv:2010.07485}\/}, .
%Type = Article
\bibitem[{Haidar et~al.(2021)Haidar, Anchuri, Rezagholizadeh, Ghaddar, Langlais
  \& Poupart}]{haidar2021rail}
\bibinfo{author}{Haidar, M.~A.}, \bibinfo{author}{Anchuri, N.},
  \bibinfo{author}{Rezagholizadeh, M.}, \bibinfo{author}{Ghaddar, A.},
  \bibinfo{author}{Langlais, P.}, \& \bibinfo{author}{Poupart, P.}
  (\bibinfo{year}{2021}).
\newblock \bibinfo{title}{Rail-kd: Random intermediate layer mapping for
  knowledge distillation}.
\newblock {\it \bibinfo{journal}{arXiv preprint arXiv:2109.10164}\/}, .
%Type = Article
\bibitem[{Han et~al.(2016)Han, Mao \& Dally}]{deepcompression2016iclr}
\bibinfo{author}{Han, S.}, \bibinfo{author}{Mao, H.}, \&
  \bibinfo{author}{Dally, W.~J.} (\bibinfo{year}{2016}).
\newblock \bibinfo{title}{Deep compression: Compressing deep neural networks
  with pruning, trained quantization and huffman coding}.
\newblock {\it \bibinfo{journal}{International Conference on Learning
  Representations (ICLR)}\/}, .
%Type = Inproceedings
\bibitem[{He et~al.(2016)He, Zhang, Ren \& Sun}]{resnet}
\bibinfo{author}{He, K.}, \bibinfo{author}{Zhang, X.}, \bibinfo{author}{Ren,
  S.}, \& \bibinfo{author}{Sun, J.} (\bibinfo{year}{2016}).
\newblock \bibinfo{title}{Deep residual learning for image recognition}.
\newblock In {\it \bibinfo{booktitle}{Proceedings of the IEEE conference on
  computer vision and pattern recognition}\/} (pp. \bibinfo{pages}{770--778}).
%Type = Inproceedings
\bibitem[{Heo et~al.(2019{\natexlab{a}})Heo, Kim, Yun, Park, Kwak \&
  Choi}]{heo2019comprehensive}
\bibinfo{author}{Heo, B.}, \bibinfo{author}{Kim, J.}, \bibinfo{author}{Yun,
  S.}, \bibinfo{author}{Park, H.}, \bibinfo{author}{Kwak, N.}, \&
  \bibinfo{author}{Choi, J.~Y.} (\bibinfo{year}{2019}{\natexlab{a}}).
\newblock \bibinfo{title}{A comprehensive overhaul of feature distillation}.
\newblock In {\it \bibinfo{booktitle}{Proceedings of the IEEE International
  Conference on Computer Vision}\/} (pp. \bibinfo{pages}{1921--1930}).
%Type = Inproceedings
\bibitem[{Heo et~al.(2019{\natexlab{b}})Heo, Lee, Yun \&
  Choi}]{heo2019knowledge}
\bibinfo{author}{Heo, B.}, \bibinfo{author}{Lee, M.}, \bibinfo{author}{Yun,
  S.}, \& \bibinfo{author}{Choi, J.~Y.} (\bibinfo{year}{2019}{\natexlab{b}}).
\newblock \bibinfo{title}{Knowledge transfer via distillation of activation
  boundaries formed by hidden neurons}.
\newblock In {\it \bibinfo{booktitle}{Proceedings of the AAAI Conference on
  Artificial Intelligence}\/} (pp. \bibinfo{pages}{3779--3787}).
\newblock volume~\bibinfo{volume}{33}.
%Type = Article
\bibitem[{Hinton et~al.(2015)Hinton, Vinyals \& Dean}]{hinton2015distilling}
\bibinfo{author}{Hinton, G.}, \bibinfo{author}{Vinyals, O.}, \&
  \bibinfo{author}{Dean, J.} (\bibinfo{year}{2015}).
\newblock \bibinfo{title}{Distilling the knowledge in a neural network}.
\newblock {\it \bibinfo{journal}{arXiv preprint arXiv:1503.02531}\/}, .
%Type = Inproceedings
\bibitem[{Howard et~al.(2019)Howard, Sandler, Chu, Chen, Chen, Tan, Wang, Zhu,
  Pang, Vasudevan et~al.}]{howard2019searching}
\bibinfo{author}{Howard, A.}, \bibinfo{author}{Sandler, M.},
  \bibinfo{author}{Chu, G.}, \bibinfo{author}{Chen, L.-C.},
  \bibinfo{author}{Chen, B.}, \bibinfo{author}{Tan, M.}, \bibinfo{author}{Wang,
  W.}, \bibinfo{author}{Zhu, Y.}, \bibinfo{author}{Pang, R.},
  \bibinfo{author}{Vasudevan, V.} et~al. (\bibinfo{year}{2019}).
\newblock \bibinfo{title}{Searching for mobilenetv3}.
\newblock In {\it \bibinfo{booktitle}{Proceedings of the IEEE/CVF International
  Conference on Computer Vision}\/} (pp. \bibinfo{pages}{1314--1324}).
%Type = Article
\bibitem[{Huang \& Wang(2017)}]{huang2017like}
\bibinfo{author}{Huang, Z.}, \& \bibinfo{author}{Wang, N.}
  (\bibinfo{year}{2017}).
\newblock \bibinfo{title}{Like what you like: Knowledge distill via neuron
  selectivity transfer}.
\newblock {\it \bibinfo{journal}{arXiv preprint arXiv:1707.01219}\/}, .
%Type = Article
\bibitem[{Huang et~al.(2022)Huang, Yang, Zhou, Li, Gong \&
  Chen}]{huang2022feature}
\bibinfo{author}{Huang, Z.}, \bibinfo{author}{Yang, S.}, \bibinfo{author}{Zhou,
  M.~C.}, \bibinfo{author}{Li, Z.}, \bibinfo{author}{Gong, Z.}, \&
  \bibinfo{author}{Chen, Y.} (\bibinfo{year}{2022}).
\newblock \bibinfo{title}{Feature map distillation of thin nets for
  low-resolution object recognition}.
\newblock {\it \bibinfo{journal}{IEEE Transactions on Image Processing}\/}, .
%Type = Article
\bibitem[{Iandola et~al.(2016)Iandola, Han, Moskewicz, Ashraf, Dally \&
  Keutzer}]{iandola2016squeezenet}
\bibinfo{author}{Iandola, F.~N.}, \bibinfo{author}{Han, S.},
  \bibinfo{author}{Moskewicz, M.~W.}, \bibinfo{author}{Ashraf, K.},
  \bibinfo{author}{Dally, W.~J.}, \& \bibinfo{author}{Keutzer, K.}
  (\bibinfo{year}{2016}).
\newblock \bibinfo{title}{Squeezenet: Alexnet-level accuracy with 50x fewer
  parameters and $<$0.5mb model size}.
\newblock {\it \bibinfo{journal}{arXiv:1602.07360}\/}, .
%Type = Misc
\bibitem[{Iqbal(2018)}]{harisIqbal}
\bibinfo{author}{Iqbal, H.} (\bibinfo{year}{2018}).
\newblock \bibinfo{title}{Harisiqbal88/plotneuralnet v1.0.0}.
\newblock \URLprefix \url{https://doi.org/10.5281/zenodo.2526396}.
  \DOIprefix\doi{10.5281/zenodo.2526396}.
%Type = Article
\bibitem[{Jain(2010)}]{k-means}
\bibinfo{author}{Jain, A.~K.} (\bibinfo{year}{2010}).
\newblock \bibinfo{title}{Data clustering: 50 years beyond k-means}.
\newblock {\it \bibinfo{journal}{Pattern recognition letters}\/},  {\it
  \bibinfo{volume}{31}\/}, \bibinfo{pages}{651--666}.
%Type = Inproceedings
\bibitem[{Jang et~al.(2019)Jang, Lee, Hwang \& Shin}]{jang2019learning}
\bibinfo{author}{Jang, Y.}, \bibinfo{author}{Lee, H.}, \bibinfo{author}{Hwang,
  S.~J.}, \& \bibinfo{author}{Shin, J.} (\bibinfo{year}{2019}).
\newblock \bibinfo{title}{Learning what and where to transfer}.
\newblock In {\it \bibinfo{booktitle}{International Conference on Machine
  Learning}\/} (pp. \bibinfo{pages}{3030--3039}).
\newblock \bibinfo{organization}{PMLR}.
%Type = Inproceedings
\bibitem[{Ji et~al.(2021)Ji, Heo \& Park}]{ji2021show}
\bibinfo{author}{Ji, M.}, \bibinfo{author}{Heo, B.}, \& \bibinfo{author}{Park,
  S.} (\bibinfo{year}{2021}).
\newblock \bibinfo{title}{Show, attend and distill: Knowledge distillation via
  attention-based feature matching}.
%Type = Article
\bibitem[{Kang \& Kang(2021)}]{kang2021data}
\bibinfo{author}{Kang, M.}, \& \bibinfo{author}{Kang, S.}
  (\bibinfo{year}{2021}).
\newblock \bibinfo{title}{Data-free knowledge distillation in neural networks
  for regression}.
\newblock {\it \bibinfo{journal}{Expert Systems with Applications}\/},  {\it
  \bibinfo{volume}{175}\/}, \bibinfo{pages}{114813}.
%Type = Article
\bibitem[{Kim et~al.(2018)Kim, Park \& Kwak}]{kim2018paraphrasing}
\bibinfo{author}{Kim, J.}, \bibinfo{author}{Park, S.}, \&
  \bibinfo{author}{Kwak, N.} (\bibinfo{year}{2018}).
\newblock \bibinfo{title}{Paraphrasing complex network: Network compression via
  factor transfer}.
\newblock {\it \bibinfo{journal}{Advances in neural information processing
  systems}\/},  {\it \bibinfo{volume}{31}\/}.
%Type = Inproceedings
\bibitem[{Kornblith et~al.(2019)Kornblith, Norouzi, Lee \&
  Hinton}]{kornblith2019similarity}
\bibinfo{author}{Kornblith, S.}, \bibinfo{author}{Norouzi, M.},
  \bibinfo{author}{Lee, H.}, \& \bibinfo{author}{Hinton, G.}
  (\bibinfo{year}{2019}).
\newblock \bibinfo{title}{Similarity of neural network representations
  revisited}.
\newblock In {\it \bibinfo{booktitle}{International Conference on Machine
  Learning}\/} (pp. \bibinfo{pages}{3519--3529}).
\newblock \bibinfo{organization}{PMLR}.
%Type = Article
\bibitem[{Krizhevsky et~al.(2017)Krizhevsky, Sutskever \& Hinton}]{alexnet}
\bibinfo{author}{Krizhevsky, A.}, \bibinfo{author}{Sutskever, I.}, \&
  \bibinfo{author}{Hinton, G.~E.} (\bibinfo{year}{2017}).
\newblock \bibinfo{title}{Imagenet classification with deep convolutional
  neural networks}.
\newblock {\it \bibinfo{journal}{Communications of the ACM}\/},  {\it
  \bibinfo{volume}{60}\/}, \bibinfo{pages}{84--90}.
%Type = Article
\bibitem[{Lan et~al.(2018)Lan, Zhu \& Gong}]{lan2018knowledge}
\bibinfo{author}{Lan, X.}, \bibinfo{author}{Zhu, X.}, \& \bibinfo{author}{Gong,
  S.} (\bibinfo{year}{2018}).
\newblock \bibinfo{title}{Knowledge distillation by on-the-fly native
  ensemble}.
\newblock {\it \bibinfo{journal}{Advances in neural information processing
  systems}\/},  {\it \bibinfo{volume}{31}\/}.
%Type = Article
\bibitem[{Li et~al.(2022{\natexlab{a}})Li, Lin, Ding, Lin, Zhuang, Huang, Ding
  \& Cao}]{li2022knowledge}
\bibinfo{author}{Li, C.}, \bibinfo{author}{Lin, M.}, \bibinfo{author}{Ding,
  Z.}, \bibinfo{author}{Lin, N.}, \bibinfo{author}{Zhuang, Y.},
  \bibinfo{author}{Huang, Y.}, \bibinfo{author}{Ding, X.}, \&
  \bibinfo{author}{Cao, L.} (\bibinfo{year}{2022}{\natexlab{a}}).
\newblock \bibinfo{title}{Knowledge condensation distillation}.
\newblock {\it \bibinfo{journal}{arXiv preprint arXiv:2207.05409}\/}, .
%Type = Article
\bibitem[{Li et~al.(2022{\natexlab{b}})Li, Lin, Wang, Wu, Tian, Shao \&
  Ji}]{li2022distilling}
\bibinfo{author}{Li, S.}, \bibinfo{author}{Lin, M.}, \bibinfo{author}{Wang,
  Y.}, \bibinfo{author}{Wu, Y.}, \bibinfo{author}{Tian, Y.},
  \bibinfo{author}{Shao, L.}, \& \bibinfo{author}{Ji, R.}
  (\bibinfo{year}{2022}{\natexlab{b}}).
\newblock \bibinfo{title}{Distilling a powerful student model via online
  knowledge distillation}.
\newblock {\it \bibinfo{journal}{IEEE Transactions on Neural Networks and
  Learning Systems}\/}, .
%Type = Inproceedings
\bibitem[{Li et~al.(2015)Li, Yosinski, Clune, Lipson \&
  Hopcroft}]{li2015convergent}
\bibinfo{author}{Li, Y.}, \bibinfo{author}{Yosinski, J.},
  \bibinfo{author}{Clune, J.}, \bibinfo{author}{Lipson, H.}, \&
  \bibinfo{author}{Hopcroft, J.~E.} (\bibinfo{year}{2015}).
\newblock \bibinfo{title}{Convergent learning: Do different neural networks
  learn the same representations?}
\newblock In {\it \bibinfo{booktitle}{FE@ NIPS}\/} (pp.
  \bibinfo{pages}{196--212}).
%Type = Inproceedings
\bibitem[{Long et~al.(2015)Long, Shelhamer \& Darrell}]{seg2012}
\bibinfo{author}{Long, J.}, \bibinfo{author}{Shelhamer, E.}, \&
  \bibinfo{author}{Darrell, T.} (\bibinfo{year}{2015}).
\newblock \bibinfo{title}{Fully convolutional networks for semantic
  segmentation}.
\newblock In {\it \bibinfo{booktitle}{Proceedings of the IEEE conference on
  computer vision and pattern recognition}\/} (pp.
  \bibinfo{pages}{3431--3440}).
%Type = Article
\bibitem[{Morcos et~al.(2018)Morcos, Raghu \& Bengio}]{morcos2018insights}
\bibinfo{author}{Morcos, A.}, \bibinfo{author}{Raghu, M.}, \&
  \bibinfo{author}{Bengio, S.} (\bibinfo{year}{2018}).
\newblock \bibinfo{title}{Insights on representational similarity in neural
  networks with canonical correlation}.
\newblock {\it \bibinfo{journal}{Advances in Neural Information Processing
  Systems}\/},  {\it \bibinfo{volume}{31}\/}, \bibinfo{pages}{5727--5736}.
%Type = Inproceedings
\bibitem[{Nam et~al.(2022)Nam, Lee, Heo \& Lee}]{nam2022improving}
\bibinfo{author}{Nam, G.}, \bibinfo{author}{Lee, H.}, \bibinfo{author}{Heo,
  B.}, \& \bibinfo{author}{Lee, J.} (\bibinfo{year}{2022}).
\newblock \bibinfo{title}{Improving ensemble distillation with weight averaging
  and diversifying perturbation}.
\newblock In {\it \bibinfo{booktitle}{International Conference on Machine
  Learning}\/} (pp. \bibinfo{pages}{16353--16367}).
\newblock \bibinfo{organization}{PMLR}.
%Type = Article
\bibitem[{Neill et~al.(2020)Neill, Steeg \& Galstyan}]{neill2020compressing}
\bibinfo{author}{Neill, J.~O.}, \bibinfo{author}{Steeg, G.~V.}, \&
  \bibinfo{author}{Galstyan, A.} (\bibinfo{year}{2020}).
\newblock \bibinfo{title}{Compressing deep neural networks via layer fusion}.
\newblock {\it \bibinfo{journal}{arXiv preprint arXiv:2007.14917}\/}, .
%Type = Article
\bibitem[{Ni et~al.(2022)Ni, Shen \& Zhao}]{ni2022federated}
\bibinfo{author}{Ni, X.}, \bibinfo{author}{Shen, X.}, \& \bibinfo{author}{Zhao,
  H.} (\bibinfo{year}{2022}).
\newblock \bibinfo{title}{Federated optimization via knowledge codistillation}.
\newblock {\it \bibinfo{journal}{Expert Systems with Applications}\/},  {\it
  \bibinfo{volume}{191}\/}, \bibinfo{pages}{116310}.
%Type = Inproceedings
\bibitem[{Park et~al.(2019)Park, Kim, Lu \& Cho}]{park2019relational}
\bibinfo{author}{Park, W.}, \bibinfo{author}{Kim, D.}, \bibinfo{author}{Lu,
  Y.}, \& \bibinfo{author}{Cho, M.} (\bibinfo{year}{2019}).
\newblock \bibinfo{title}{Relational knowledge distillation}.
\newblock In {\it \bibinfo{booktitle}{Proceedings of the IEEE/CVF Conference on
  Computer Vision and Pattern Recognition}\/} (pp.
  \bibinfo{pages}{3967--3976}).
%Type = Article
\bibitem[{Parkhi et~al.(2015)Parkhi, Vedaldi \& Zisserman}]{facerecognition}
\bibinfo{author}{Parkhi, O.~M.}, \bibinfo{author}{Vedaldi, A.}, \&
  \bibinfo{author}{Zisserman, A.} (\bibinfo{year}{2015}).
\newblock \bibinfo{title}{Deep face recognition}, .
%Type = Inproceedings
\bibitem[{Passalis \& Tefas(2018)}]{passalis2018learning}
\bibinfo{author}{Passalis, N.}, \& \bibinfo{author}{Tefas, A.}
  (\bibinfo{year}{2018}).
\newblock \bibinfo{title}{Learning deep representations with probabilistic
  knowledge transfer}.
\newblock In {\it \bibinfo{booktitle}{Proceedings of the European Conference on
  Computer Vision (ECCV)}\/} (pp. \bibinfo{pages}{268--284}).
%Type = Inproceedings
\bibitem[{Passban et~al.(2021)Passban, Wu, Rezagholizadeh \&
  Liu}]{passban2021alp}
\bibinfo{author}{Passban, P.}, \bibinfo{author}{Wu, Y.},
  \bibinfo{author}{Rezagholizadeh, M.}, \& \bibinfo{author}{Liu, Q.}
  (\bibinfo{year}{2021}).
\newblock \bibinfo{title}{Alp-kd: Attention-based layer projection for
  knowledge distillation}.
\newblock In {\it \bibinfo{booktitle}{Proceedings of the AAAI Conference on
  Artificial Intelligence}\/} (pp. \bibinfo{pages}{13657--13665}).
\newblock volume~\bibinfo{volume}{35}.
%Type = Inproceedings
\bibitem[{Peng et~al.(2019)Peng, Jin, Liu, Li, Wu, Liu, Zhou \&
  Zhang}]{peng2019correlation}
\bibinfo{author}{Peng, B.}, \bibinfo{author}{Jin, X.}, \bibinfo{author}{Liu,
  J.}, \bibinfo{author}{Li, D.}, \bibinfo{author}{Wu, Y.},
  \bibinfo{author}{Liu, Y.}, \bibinfo{author}{Zhou, S.}, \&
  \bibinfo{author}{Zhang, Z.} (\bibinfo{year}{2019}).
\newblock \bibinfo{title}{Correlation congruence for knowledge distillation}.
\newblock In {\it \bibinfo{booktitle}{Proceedings of the IEEE/CVF International
  Conference on Computer Vision}\/} (pp. \bibinfo{pages}{5007--5016}).
%Type = Inproceedings
\bibitem[{Raghu et~al.(2017)Raghu, Gilmer, Yosinski \&
  Sohl-Dickstein}]{raghu2017svcca}
\bibinfo{author}{Raghu, M.}, \bibinfo{author}{Gilmer, J.},
  \bibinfo{author}{Yosinski, J.}, \& \bibinfo{author}{Sohl-Dickstein, J.}
  (\bibinfo{year}{2017}).
\newblock \bibinfo{title}{Svcca: Singular vector canonical correlation analysis
  for deep learning dynamics and interpretability}.
\newblock In {\it \bibinfo{booktitle}{Advances in neural information processing
  systems}\/} (pp. \bibinfo{pages}{6076--6085}).
%Type = Article
\bibitem[{Ramsay et~al.(1984)Ramsay, ten Berge \& Styan}]{ramsay1984matrix}
\bibinfo{author}{Ramsay, J.}, \bibinfo{author}{ten Berge, J.}, \&
  \bibinfo{author}{Styan, G.} (\bibinfo{year}{1984}).
\newblock \bibinfo{title}{Matrix correlation}.
\newblock {\it \bibinfo{journal}{Psychometrika}\/},  {\it
  \bibinfo{volume}{49}\/}, \bibinfo{pages}{403--423}.
%Type = Inproceedings
\bibitem[{{Rao} \& {Frtunikj}(2018)}]{self_driving2018}
\bibinfo{author}{{Rao}, Q.}, \& \bibinfo{author}{{Frtunikj}, J.}
  (\bibinfo{year}{2018}).
\newblock \bibinfo{title}{Deep learning for self-driving cars: Chances and
  challenges}.
\newblock In {\it \bibinfo{booktitle}{2018 IEEE/ACM 1st International Workshop
  on Software Engineering for AI in Autonomous Systems (SEFAIAS)}\/} (pp.
  \bibinfo{pages}{35--38}).
%Type = Article
\bibitem[{Redmon \& Farhadi(2018)}]{yolov3}
\bibinfo{author}{Redmon, J.}, \& \bibinfo{author}{Farhadi, A.}
  (\bibinfo{year}{2018}).
\newblock \bibinfo{title}{Yolov3: An incremental improvement}.
\newblock {\it \bibinfo{journal}{arXiv preprint arXiv:1804.02767}\/}, .
%Type = Inproceedings
\bibitem[{Ren et~al.(2022)Ren, Gao, Hua, Xue, Tian, He \& Zhao}]{ren2022co}
\bibinfo{author}{Ren, S.}, \bibinfo{author}{Gao, Z.}, \bibinfo{author}{Hua,
  T.}, \bibinfo{author}{Xue, Z.}, \bibinfo{author}{Tian, Y.},
  \bibinfo{author}{He, S.}, \& \bibinfo{author}{Zhao, H.}
  (\bibinfo{year}{2022}).
\newblock \bibinfo{title}{Co-advise: Cross inductive bias distillation}.
\newblock In {\it \bibinfo{booktitle}{Proceedings of the IEEE/CVF Conference on
  Computer Vision and Pattern Recognition}\/} (pp.
  \bibinfo{pages}{16773--16782}).
%Type = Article
\bibitem[{Romero et~al.(2015)Romero, Ballas, Kahou, Chassang, Gatta \&
  Bengio}]{romero2015fitnets}
\bibinfo{author}{Romero, A.}, \bibinfo{author}{Ballas, N.},
  \bibinfo{author}{Kahou, S.~E.}, \bibinfo{author}{Chassang, A.},
  \bibinfo{author}{Gatta, C.}, \& \bibinfo{author}{Bengio, Y.}
  (\bibinfo{year}{2015}).
\newblock \bibinfo{title}{Fitnets: Hints for thin deep nets}.
\newblock {\it \bibinfo{journal}{Proc. ICLR}\/},  {\it \bibinfo{volume}{2}\/}.
%Type = Article
\bibitem[{Ruffy \& Chahal(2019)}]{ruffy2019state}
\bibinfo{author}{Ruffy, F.}, \& \bibinfo{author}{Chahal, K.}
  (\bibinfo{year}{2019}).
\newblock \bibinfo{title}{The state of knowledge distillation for
  classification}.
\newblock {\it \bibinfo{journal}{arXiv preprint arXiv:1912.10850}\/}, .
%Type = Article
\bibitem[{Russakovsky et~al.(2015)Russakovsky, Deng, Su, Krause, Satheesh, Ma,
  Huang, Karpathy, Khosla, Bernstein, Berg \& Fei-Fei}]{ILSVRC15}
\bibinfo{author}{Russakovsky, O.}, \bibinfo{author}{Deng, J.},
  \bibinfo{author}{Su, H.}, \bibinfo{author}{Krause, J.},
  \bibinfo{author}{Satheesh, S.}, \bibinfo{author}{Ma, S.},
  \bibinfo{author}{Huang, Z.}, \bibinfo{author}{Karpathy, A.},
  \bibinfo{author}{Khosla, A.}, \bibinfo{author}{Bernstein, M.},
  \bibinfo{author}{Berg, A.~C.}, \& \bibinfo{author}{Fei-Fei, L.}
  (\bibinfo{year}{2015}).
\newblock \bibinfo{title}{{ImageNet Large Scale Visual Recognition Challenge}}.
\newblock {\it \bibinfo{journal}{International Journal of Computer Vision
  (IJCV)}\/},  {\it \bibinfo{volume}{115}\/}, \bibinfo{pages}{211--252}.
  \DOIprefix\doi{10.1007/s11263-015-0816-y}.
%Type = Inproceedings
\bibitem[{Sandler et~al.(2018)Sandler, Howard, Zhu, Zhmoginov \&
  Chen}]{sandler2018mobilenetv2}
\bibinfo{author}{Sandler, M.}, \bibinfo{author}{Howard, A.},
  \bibinfo{author}{Zhu, M.}, \bibinfo{author}{Zhmoginov, A.}, \&
  \bibinfo{author}{Chen, L.-C.} (\bibinfo{year}{2018}).
\newblock \bibinfo{title}{Mobilenetv2: Inverted residuals and linear
  bottlenecks}.
\newblock In {\it \bibinfo{booktitle}{Proceedings of the IEEE conference on
  computer vision and pattern recognition}\/} (pp.
  \bibinfo{pages}{4510--4520}).
%Type = Article
\bibitem[{Shao \& Chen(2021)}]{shao2021multi}
\bibinfo{author}{Shao, B.}, \& \bibinfo{author}{Chen, Y.}
  (\bibinfo{year}{2021}).
\newblock \bibinfo{title}{Multi-granularity for knowledge distillation}.
\newblock {\it \bibinfo{journal}{Image and Vision Computing}\/},  {\it
  \bibinfo{volume}{115}\/}, \bibinfo{pages}{104286}.
%Type = Article
\bibitem[{Simonyan \& Zisserman(2014)}]{VGG}
\bibinfo{author}{Simonyan, K.}, \& \bibinfo{author}{Zisserman, A.}
  (\bibinfo{year}{2014}).
\newblock \bibinfo{title}{Very deep convolutional networks for large-scale
  image recognition}.
\newblock {\it \bibinfo{journal}{arXiv preprint arXiv:1409.1556}\/}, .
%Type = Inproceedings
\bibitem[{Sun et~al.(2019)Sun, Cheng, Gan \& Liu}]{sun2019patient}
\bibinfo{author}{Sun, S.}, \bibinfo{author}{Cheng, Y.}, \bibinfo{author}{Gan,
  Z.}, \& \bibinfo{author}{Liu, J.} (\bibinfo{year}{2019}).
\newblock \bibinfo{title}{Patient knowledge distillation for bert model
  compression}.
\newblock In {\it \bibinfo{booktitle}{Proceedings of the 2019 Conference on
  Empirical Methods in Natural Language Processing and the 9th International
  Joint Conference on Natural Language Processing (EMNLP-IJCNLP)}\/} (pp.
  \bibinfo{pages}{4323--4332}).
%Type = Article
\bibitem[{Sze et~al.(2017)Sze, Chen, Yang \& Emer}]{sze2017efficient}
\bibinfo{author}{Sze, V.}, \bibinfo{author}{Chen, Y.-H.},
  \bibinfo{author}{Yang, T.-J.}, \& \bibinfo{author}{Emer, J.~S.}
  (\bibinfo{year}{2017}).
\newblock \bibinfo{title}{Efficient processing of deep neural networks: A
  tutorial and survey}.
\newblock {\it \bibinfo{journal}{Proceedings of the IEEE}\/},  {\it
  \bibinfo{volume}{105}\/}, \bibinfo{pages}{2295--2329}.
%Type = Inproceedings
\bibitem[{Szegedy et~al.(2017)Szegedy, Ioffe, Vanhoucke \&
  Alemi}]{szegedy2017inception}
\bibinfo{author}{Szegedy, C.}, \bibinfo{author}{Ioffe, S.},
  \bibinfo{author}{Vanhoucke, V.}, \& \bibinfo{author}{Alemi, A.~A.}
  (\bibinfo{year}{2017}).
\newblock \bibinfo{title}{Inception-v4, inception-resnet and the impact of
  residual connections on learning}.
\newblock In {\it \bibinfo{booktitle}{Thirty-first AAAI conference on
  artificial intelligence}\/}.
%Type = Inproceedings
\bibitem[{Tian et~al.(2019)Tian, Krishnan \& Isola}]{tian2019contrastive}
\bibinfo{author}{Tian, Y.}, \bibinfo{author}{Krishnan, D.}, \&
  \bibinfo{author}{Isola, P.} (\bibinfo{year}{2019}).
\newblock \bibinfo{title}{Contrastive representation distillation}.
\newblock In {\it \bibinfo{booktitle}{International Conference on Learning
  Representations}\/}.
%Type = Inproceedings
\bibitem[{Tung \& Mori(2019)}]{tung2019similarity}
\bibinfo{author}{Tung, F.}, \& \bibinfo{author}{Mori, G.}
  (\bibinfo{year}{2019}).
\newblock \bibinfo{title}{Similarity-preserving knowledge distillation}.
\newblock In {\it \bibinfo{booktitle}{Proceedings of the IEEE/CVF International
  Conference on Computer Vision}\/} (pp. \bibinfo{pages}{1365--1374}).
%Type = Article
\bibitem[{Wang \& Yoon(2021)}]{wang2021knowledge}
\bibinfo{author}{Wang, L.}, \& \bibinfo{author}{Yoon, K.-J.}
  (\bibinfo{year}{2021}).
\newblock \bibinfo{title}{Knowledge distillation and student-teacher learning
  for visual intelligence: A review and new outlooks}.
\newblock {\it \bibinfo{journal}{IEEE Transactions on Pattern Analysis and
  Machine Intelligence}\/}, .
%Type = Inproceedings
\bibitem[{Wu \& Deng(2022)}]{wu2022single}
\bibinfo{author}{Wu, A.}, \& \bibinfo{author}{Deng, C.} (\bibinfo{year}{2022}).
\newblock \bibinfo{title}{Single-domain generalized object detection in urban
  scene via cyclic-disentangled self-distillation}.
\newblock In {\it \bibinfo{booktitle}{Proceedings of the IEEE/CVF Conference on
  Computer Vision and Pattern Recognition}\/} (pp. \bibinfo{pages}{847--856}).
%Type = Article
\bibitem[{Yang et~al.(2022)Yang, Zhou, Zhang, Sun, Wu, Wang \&
  Ye}]{yang2022multi}
\bibinfo{author}{Yang, D.}, \bibinfo{author}{Zhou, Y.}, \bibinfo{author}{Zhang,
  A.}, \bibinfo{author}{Sun, X.}, \bibinfo{author}{Wu, D.},
  \bibinfo{author}{Wang, W.}, \& \bibinfo{author}{Ye, Q.}
  (\bibinfo{year}{2022}).
\newblock \bibinfo{title}{Multi-view correlation distillation for incremental
  object detection}.
\newblock {\it \bibinfo{journal}{Pattern Recognition}\/},  {\it
  \bibinfo{volume}{131}\/}, \bibinfo{pages}{108863}.
%Type = Inproceedings
\bibitem[{Yim et~al.(2017)Yim, Joo, Bae \& Kim}]{yim2017gift}
\bibinfo{author}{Yim, J.}, \bibinfo{author}{Joo, D.}, \bibinfo{author}{Bae,
  J.}, \& \bibinfo{author}{Kim, J.} (\bibinfo{year}{2017}).
\newblock \bibinfo{title}{A gift from knowledge distillation: Fast
  optimization, network minimization and transfer learning}.
\newblock In {\it \bibinfo{booktitle}{Proceedings of the IEEE Conference on
  Computer Vision and Pattern Recognition}\/} (pp.
  \bibinfo{pages}{4133--4141}).
%Type = Article
\bibitem[{Yosinski et~al.(2014)Yosinski, Clune, Bengio \&
  Lipson}]{yosinski2014transferable}
\bibinfo{author}{Yosinski, J.}, \bibinfo{author}{Clune, J.},
  \bibinfo{author}{Bengio, Y.}, \& \bibinfo{author}{Lipson, H.}
  (\bibinfo{year}{2014}).
\newblock \bibinfo{title}{How transferable are features in deep neural
  networks?}
\newblock {\it \bibinfo{journal}{Advances in neural information processing
  systems}\/},  {\it \bibinfo{volume}{27}\/}.
%Type = Inproceedings
\bibitem[{Zagoruyko \& Komodakis(2016)}]{zagoruyko2016wide}
\bibinfo{author}{Zagoruyko, S.}, \& \bibinfo{author}{Komodakis, N.}
  (\bibinfo{year}{2016}).
\newblock \bibinfo{title}{Wide residual networks}.
\newblock In {\it \bibinfo{booktitle}{British Machine Vision Conference
  2016}\/}.
\newblock \bibinfo{organization}{British Machine Vision Association}.
%Type = Inproceedings
\bibitem[{Zagoruyko \& Komodakis(2017)}]{Zagoruyko2017AT}
\bibinfo{author}{Zagoruyko, S.}, \& \bibinfo{author}{Komodakis, N.}
  (\bibinfo{year}{2017}).
\newblock \bibinfo{title}{Paying more attention to attention: Improving the
  performance of convolutional neural networks via attention transfer}.
\newblock In {\it \bibinfo{booktitle}{ICLR}\/}.
\newblock \URLprefix \url{https://arxiv.org/abs/1612.03928}.
%Type = Article
\bibitem[{Zhang et~al.(2021{\natexlab{a}})Zhang, Zhanga, Li, Zeng \&
  Ge}]{zhang2021student}
\bibinfo{author}{Zhang, K.}, \bibinfo{author}{Zhanga, C.}, \bibinfo{author}{Li,
  S.}, \bibinfo{author}{Zeng, D.}, \& \bibinfo{author}{Ge, S.}
  (\bibinfo{year}{2021}{\natexlab{a}}).
\newblock \bibinfo{title}{Student network learning via evolutionary knowledge
  distillation}.
\newblock {\it \bibinfo{journal}{IEEE Transactions on Circuits and Systems for
  Video Technology}\/}, .
%Type = Article
\bibitem[{Zhang et~al.(2021{\natexlab{b}})Zhang, Bao \& Ma}]{zhang2021self}
\bibinfo{author}{Zhang, L.}, \bibinfo{author}{Bao, C.}, \& \bibinfo{author}{Ma,
  K.} (\bibinfo{year}{2021}{\natexlab{b}}).
\newblock \bibinfo{title}{Self-distillation: Towards efficient and compact
  neural networks}.
\newblock {\it \bibinfo{journal}{IEEE Transactions on Pattern Analysis and
  Machine Intelligence}\/},  {\it \bibinfo{volume}{44}\/},
  \bibinfo{pages}{4388--4403}.
%Type = Inproceedings
\bibitem[{Zhang et~al.(2018)Zhang, Zhou, Lin \& Sun}]{zhang2018shufflenet}
\bibinfo{author}{Zhang, X.}, \bibinfo{author}{Zhou, X.}, \bibinfo{author}{Lin,
  M.}, \& \bibinfo{author}{Sun, J.} (\bibinfo{year}{2018}).
\newblock \bibinfo{title}{Shufflenet: An extremely efficient convolutional
  neural network for mobile devices}.
\newblock In {\it \bibinfo{booktitle}{Proceedings of the IEEE conference on
  computer vision and pattern recognition}\/} (pp.
  \bibinfo{pages}{6848--6856}).
%Type = Inproceedings
\bibitem[{Zhao et~al.(2022)Zhao, Cui, Song, Qiu \& Liang}]{zhao2022decoupled}
\bibinfo{author}{Zhao, B.}, \bibinfo{author}{Cui, Q.}, \bibinfo{author}{Song,
  R.}, \bibinfo{author}{Qiu, Y.}, \& \bibinfo{author}{Liang, J.}
  (\bibinfo{year}{2022}).
\newblock \bibinfo{title}{Decoupled knowledge distillation}.
\newblock In {\it \bibinfo{booktitle}{Proceedings of the IEEE/CVF Conference on
  Computer Vision and Pattern Recognition}\/} (pp.
  \bibinfo{pages}{11953--11962}).
%Type = Inproceedings
\bibitem[{Zhou et~al.(2021)Zhou, Song, Chen, Zhou, Wang, Yuan \&
  Zhang}]{zhou2021rethinking}
\bibinfo{author}{Zhou, H.}, \bibinfo{author}{Song, L.}, \bibinfo{author}{Chen,
  J.}, \bibinfo{author}{Zhou, Y.}, \bibinfo{author}{Wang, G.},
  \bibinfo{author}{Yuan, J.}, \& \bibinfo{author}{Zhang, Q.}
  (\bibinfo{year}{2021}).
\newblock \bibinfo{title}{Rethinking soft labels for knowledge distillation: A
  bias-variance tradeoff perspective}.
\newblock In {\it \bibinfo{booktitle}{International Conference on Learning
  Representations (ICLR)}\/}.

\end{thebibliography}

\end{document}